\documentclass[10pt,twocolumn,letterpaper]{article}

\usepackage{cvpr}              %

\usepackage{times}
\usepackage{epsfig}
\usepackage{graphicx}
\usepackage{amsmath}
\usepackage{amssymb}
\usepackage{booktabs}
\usepackage{siunitx}
\usepackage{pgfplots}
\pgfplotsset{compat=1.18}
\usepackage{multirow}
\usepackage[normalem]{ulem}
\usepackage{xpatch}
\usepackage{bbm}
\usepackage{soul}
\usepackage{appendix}

\usepackage[accsupp]{axessibility}
\usepackage{placeins}

\definecolor{cvprblue}{rgb}{0.21,0.49,0.74}
\usepackage[pagebackref,breaklinks,colorlinks,citecolor=cvprblue]{hyperref}

\def\paperID{5193} %
\def\confName{CVPR}
\def\confYear{2024}

\newcommand{\ourmodel}{\textsc{UnO}}
\newcommand{\CMU}{\textsc{4D-Occ}}
\newcommand{\freespacerendering}{\textsc{Free-Space-Rendering}}
\newcommand{\depthrendering}{\textsc{Depth-Rendering}}
\newcommand{\unbalancedourmodel}{\textsc{Unbalanced UnO}}

\newcommand{\ben}[1]{\textcolor{blue}{Ben: #1}}
\newcommand{\sergio}[1]{\textcolor{cyan}{Sergio: #1}}
\newcommand{\quin}[1]{\textcolor{purple}{Quin: #1}}
\newcommand{\thomas}[1]{\textcolor{green}{Thomas: #1}}
\newcommand{\raquel}[1]{\textcolor{red}{Raquel: #1}}

\renewcommand{\ben}[1]{}
\renewcommand{\sergio}[1]{}
\renewcommand{\quin}[1]{}
\renewcommand{\thomas}[1]{}
\renewcommand{\raquel}[1]{}

\title{UnO: Unsupervised Occupancy Fields for Perception and Forecasting}

\author{
\textbf{Ben Agro\footnotemark[1]$^{\text{ }\;1,2}$, Quinlan Sykora$^*$$^{1,2}$, Sergio Casas$^*$$^{1,2}$, Thomas Gilles$^1$, Raquel Urtasun$^{1,2}$} \\
Waabi$^1$, University of Toronto$^2$ \\
\texttt{\{bagro, qsykora, sergio, tgilles, urtasun\}@waabi.ai}
\vspace{-0.5em}
}

\begin{document}

\makeatletter
\xpatchcmd{\paragraph}{3.25ex \@plus1ex \@minus.2ex}{1.0ex plus 0.1ex minus 0.1ex}{\typeout{success!}}{\typeout{failure!}}
\makeatother

\twocolumn[{%
\renewcommand\twocolumn[1][]{#1}%
\maketitle
    \input{figures/hook.tex}
	\vspace{6mm}
}]

\begin{abstract}
\vspace{-1ex}
Perceiving the world and forecasting its future state is a critical task for self-driving. Supervised approaches leverage annotated object labels to learn a model of the world --- traditionally with object detections and trajectory predictions, or temporal bird's-eye-view (BEV) occupancy fields. However, these annotations are expensive and typically limited to a set of predefined categories that do not cover everything we might encounter on the road. Instead, we learn to perceive and forecast a continuous 4D (spatio-temporal) occupancy field with self-supervision from LiDAR data. This unsupervised world model can be easily and effectively transferred to downstream tasks. We tackle point cloud forecasting by adding a lightweight learned renderer and achieve state-of-the-art performance in Argoverse 2, nuScenes, and KITTI. To further showcase its transferability, we fine-tune our model for BEV semantic occupancy forecasting and show that it outperforms the fully supervised state-of-the-art, especially when labeled data is scarce. Finally, when compared to prior state-of-the-art on spatio-temporal geometric occupancy prediction, our 4D world model achieves a much higher recall of objects from classes relevant to self-driving.
For more information, visit the project website: \href{https://waabi.ai/research/uno}{https://waabi.ai/research/uno}.
\vspace{-1em}
\end{abstract}

\section{Introduction} \label{sec:intro}

For a self-driving vehicle (SDV) to plan its actions effectively and safely, it must be able to perceive the environment and forecast how it will evolve in the future. 
Two paradigms have been developed in order to perform these two tasks. 
The most common approach is to detect a discrete set of objects in the scene, then forecast possible future trajectories of each object \cite{casas2018intentnet,Liang_2020_CVPR,weng2020ptp,weng20203d,salzmann2020trajectron,ivanovic2018the,ivanovic2020multimodal,gu2021densetnt,cui2021lookout}.
More recently, bird's-eye view (BEV) \textit{semantic occupancy fields} \cite{mahjourian2022occupancy,casas2021mp3,sadat2020perceive,philion2020lift,hu2021fiery,agro2023implicit} have become popular as they avoid thresholding confidence scores and better represent uncertainty about future motion.

\renewcommand{\thefootnote}{\fnsymbol{footnote}}
\footnotetext[1]{Denotes equal contribution.}
\renewcommand{\thefootnote}{\arabic{footnote}}

These approaches leverage supervision from human annotations to learn a model of the world.
Unfortunately, their performance is bounded by the scale and expressiveness of the human annotations. 
Due to the high cost of these labels, the amount of available labeled data is orders of magnitude smaller than the amount of unlabeled data.
Furthermore, these labels are typically restricted to a predefined set of object classes and the object shape is approximated with a 3D bounding box, which for many classes is a very crude approximation. 
Thus, rare events and infrequent objects are seldom included in labeled data, limiting the safety of current self-driving systems. 

This motivates the development of methods that can leverage vast amounts of unlabeled sensor data to learn representations of the world. 
Prior works proposed to directly predict future point clouds from past point clouds \cite{mersch2022self, weng2022s2net, weng2021inverting, wilson2023argoverse}. 
However, this makes the task unnecessarily difficult, as the model must learn not only a model of the world, but also the sensor extrinsics and intrinsics as well as LiDAR properties such as ray reflectivity, which is a complex function of materials and incidence angle. 
To address this issue, \CMU{}~\cite{khurana2023point} proposed to learn future \textit{geometric occupancy voxel grids} exploiting the known sensor intrinsics and extrinsics. 
However, this method is limited by the use of a quantized voxel grid and a LiDAR depth-rendering objective that optimizes for optical density via regression. 
As shown in our experiments, this results in models that struggle learning the dynamics of the world.
Furthermore, whether the learned representations are useful for downstream tasks other than point cloud forecasting remains unknown.

Our goal is to learn a model of the world that can exploit large-scale unlabeled LiDAR data and can be easily and effectively transferred to perform downstream perception and forecasting tasks. 
Towards this goal, we propose a novel unsupervised task: forecasting continuous 4D (3D space and time) occupancy fields (\cref{fig:hook}\textcolor{red}{.a}) from LiDAR observations. 
This objective is suitable for learning general representations because accurately predicting spatio-temporal occupancy fields requires an understanding of the world's \textit{geometry} (e.g., to predict shapes of partially occluded objects), \textit{dynamics} (e.g., to predict where moving objects will be in the future) and \textit{semantics} (e.g., to understand the rules of the road). 
Importantly, we employ an implicit architecture to allow 
our model to be queried at any given continuous point $(x, y, z, t)$ in space and future time. 
Our world model, which we dub \ourmodel{} (\textsc{Un}supervised \textsc{O}ccupancy), learns common sense concepts such as the full extent of objects, 
even though the input LiDAR only sees a portion of the object.
The ability to forecast multi-modal futures with associated uncertainty also emerges; e.g., \ourmodel{} can predict that a vehicle may or may not lane change, and a pedestrian may stay on the sidewalk or enter the crosswalk. 

To demonstrate the generalizability and effectiveness of our world model, we show that it can be transferred to two important downstream tasks: point cloud forecasting (\cref{fig:hook}\textcolor{red}{.b}) and supervised BEV semantic occupancy prediction (\cref{fig:hook}\textcolor{red}{.c}). 
For point cloud forecasting, \ourmodel{} surpasses the state-of-the-art in Argoverse 2, nuScenes, and KITTI by learning a simple ray depth renderer on top of the occupancy predictions.
For BEV semantic occupancy prediction, we show that fine-tuning \ourmodel{} outperforms fully supervised approaches, and that the improvement is particularly large when limited labels are available for training, demonstrating impressive few-shot generalization.

\section{Related Work}

\paragraph{Point Cloud Forecasting:}

Predicting future point clouds from past observations provides a setting to learn world models from large unlabeled datasets.
Prior works aim to predict point clouds directly
\cite{mersch2022self, weng2021inverting, weng2022s2net,wilson2023argoverse},
which requires forecasting the sensor extrinsics, including where the ego vehicle will be in the future; the sensor intrinsics, e.g., the sampling pattern specific to the LiDAR sensor; and the shape and motion of objects in the scene.
\CMU{} \cite{khurana2023point} reformulates the point cloud prediction
task to factor out the sensor extrinsics and intrinsics.
Specifically, it first predicts occupancy in space and future time as a 4D voxel grid and then predicts the depth of LiDAR rays given by the sensor intrinsics and future extrinsics through a NERF-like rendering of the occupancy. 
However, \CMU{} does not attain all the desired capabilities of a world model. For instance, it struggles to forecast moving objects. 
We believe this is because their regression loss emphasizes 
accurate depth predictions instead of occupancy classification, and their 4D voxel grids introduce quantization errors.
While our work also builds on the idea of using an internal 4D geometric occupancy representation, we improve upon prior art by learning a continuous 4D occupancy field through a classification objective instead of a voxel grid through regression, and by decoupling the training of the occupancy (i.e., world model) and the point cloud renderer (i.e., downstream task).

\paragraph{Occupancy Forecasting from Sensors:}
This task consists of predicting the probability that a certain area of space will be occupied in the future, directly from sensory data like LiDAR \cite{casas2021mp3,sadat2020perceive} and camera \cite{hu2021fiery, hu2022st, philion2020lift, exquisite}.
The occupancy predictions are often \textit{semantic}; they represent
classes of objects (e.g., vehicle, cyclist) relevant to downstream tasks like motion planning for SDVs.
Because most planners \cite{fan2018baidu,cui2021lookout,casas2021mp3,renz2022plant} for autonomous driving reason in 2D BEV space, prior work has focused on forecasting accurate occupancy in BEV.
P3 \cite{sadat2020perceive}, MP3 \cite{casas2021mp3} and FIERY \cite{hu2021fiery} predict temporal BEV semantic occupancy grids with convolutional neural networks directly from sensor data. 
UniAD \cite{exquisite} predicts both trajectory predictions and BEV semantic occupancy forecasts.
P3, MP3 and UniAD all demonstrate that utilizing BEV semantic occupancy can improve motion planning for self-driving.
ImplicitO \cite{agro2023implicit} introduces an attention-based architecture for predicting occupancy at any spatio-temporal continuous point $(x, y, t)$ in BEV, improving efficiency by 
only producing occupancy at user specified query points.
Other works have sought to predict BEV occupancy without semantic labels. For instance, \cite{khurana2022differentiable} leverages self-supervision from visibility maps generated with future LiDAR point clouds, thereby combining the occupancy of semantic classes and the background, which 
we refer to as \textit{geometric occupancy}. 

\paragraph{Pre-Training for Perception:}

The ultimate goal of many representation learning methods is to improve performance in downstream vision tasks like 2D/3D object detection and semantic segmentation.
Masked AutoEncoders (MAEs) \cite{he2022masked} have recently gained traction: mask random image patches and use their pixel values as reconstruction targets.
VoxelMAE \cite{hess2023masked} extends this concept to LiDAR data by voxelizing the point cloud data
and demonstrates this is effective as pre-training for object detection in self-driving.
UniPAD \cite{yang2023unipad} builds on these works by learning a NeRF-like world model to reconstruct color and depth data from masked multi-modal image and point cloud inputs.
ALSO \cite{boulch2023also} shows that surface reconstruction from present-time LiDAR rays can be a strong pre-training task. 
In contrast to these prior works, we focus not only on pre-training for understanding the present-time world state from sensor data, but also to forecast the future.

\vspace{-1ex}
\section{Unsupervised Occupancy World Model} \label{sec:method-UnO}
\vspace{-1ex}

Time-of-flight LiDARs measure the distance of a surface from the sensor, which we refer to as \textit{depth}, by emitting rays of light from a pulsed laser.
A LiDAR point occurs when the ray hits an object and the laser returns.
The depth is calculated based on the time it took for the ray to come back to the sensor.
Thus, a LiDAR point indicates that the space along the ray between the sensor and the observed point is unoccupied, and that there is some occupied space directly after that point in the ray direction.

Although LiDAR point clouds are usually displayed as a 360$^\circ$ scan called a \textit{sweep}, the points are actually acquired at different times.
For instance, in mechanical spinning LiDARs, different parts of the scene are scanned as a set of beams emit rays into a polygon mirror while it rotates \cite{raj2020survey}.
Accounting for the right time of emission is particularly important for objects moving at high speeds, and also if the data collection platform is equipped with multiple LiDAR sensors that scan the scene asynchronously, which is the case in many modern SDVs. 

We take inspiration from these intuitions and propose an unsupervised task (\cref{sec:method-selfsupervision}) as well as a simple yet effective model (\cref{sec:method-model}) to learn the world's geometry, dynamics and even semantics from future point clouds. %

\subsection{Unsupervised Task} \label{sec:method-selfsupervision}

We assume that we have access to a calibrated and localized vehicle as our data collection platform. Note that these assumptions are not restrictive as this is the norm for self-driving platforms. This implies that we know where the sensors are located on the vehicle, how the vehicle moves around the scene, and we can capture the observed LiDAR points over time. 
We can then make use of the known sensor intrinsics and
extrinsics to construct occupancy pseudo-labels over space and time providing us with positive (occupied) and negative (unoccupied) supervision
for every emitted ray that returned, as shown in \cref{fig:objective}. 
Note that we have partial supervision as there are regions where we do not know the occupancy due to occlusion or scene properties such non-reflective materials.

\definecolor{mypink1}{RGB}{156, 0, 252}
\begin{figure}[t]
    \centering
    \includegraphics[width=1.0\linewidth,trim={0cm 0cm 0cm 0cm},clip]{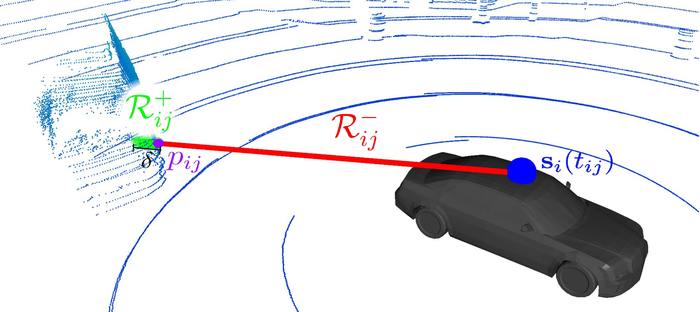}
    \caption{
        \ourmodel{}'s occupancy pseudo-labels: 
        a laser beam emitted from sensor position $\textcolor{blue}{s_i}$ at time $\textcolor{blue}{t_{ij}}$ 
        returns the point $\textcolor{mypink1}{p_{ij}}$, meaning that the ray segment $\textcolor{red}{\mathcal{R}_{ij}^{-}}$ 
        is unoccupied space and the segment within a buffer $\delta$ after the lidar return is occupied space $\textcolor{green}{\mathcal{R}_{ij}^{+}}$.
      }
    \label{fig:objective}
    \vspace{-1em}
\end{figure}

More formally, let $\mathbf{s}_i(t) = (s_{i}^x(t), s_{i}^y(t), s_{i}^z(t))$ be the 3D position at time $t$ of the $i$-th LiDAR sensor 
mounted on the SDV. Let $\mathbf{p}_{ij} = (p_{ij}^x, p_{ij}^y, p_{ij}^z)$ be the $j$-th LiDAR
point returned by sensor $i$, with emission time denoted $t_{ij}$.
For a single LiDAR point, the line segment between $\mathbf{s}_i(t_{ij})$ and $\mathbf{p}_{ij}$ is unoccupied
\begin{align}
    \mathcal{R}_{ij}^- = \{ \mathbf{s}_i(t_{ij}) + (\mathbf{p}_{ij} - \mathbf{s}_i(t_{ij})) r \; | \; \forall r \in (0, 1) \}.
\end{align}
We also produce a positive occupancy directly behind the LiDAR point $\mathbf{p}_{ij}$:
\begin{align}
    \mathcal{R}_{ij}^+ = \{ \mathbf{p}_{ij} + \frac{ (\mathbf{p}_{ij} - \mathbf{s}_i(t_{ij})) }{ || \mathbf{p}_{ij} - \mathbf{s}_i(t_{ij}) ||_2 } r \; | \; \forall r \in [0, \delta] \},
\end{align}
where $\delta$ is a small buffer region, see \cref{fig:objective} (e.g., we use $\delta = \SI{0.1}{m}$ for all of our experiments).
Note that $\delta$ should be small; if it were too large then the occupied regions would misrepresent the true shape of objects (especially thin structures), 
and multiple rays hitting the same object could conflict on their assignment of occupied / unoccupied space. 

\subsection{Unsupervised Occupancy Model (\ourmodel{})} \label{sec:method-model}

\begin{figure*}[t]
    \centering
    \vspace{-15pt}
    \includegraphics[width=\linewidth]{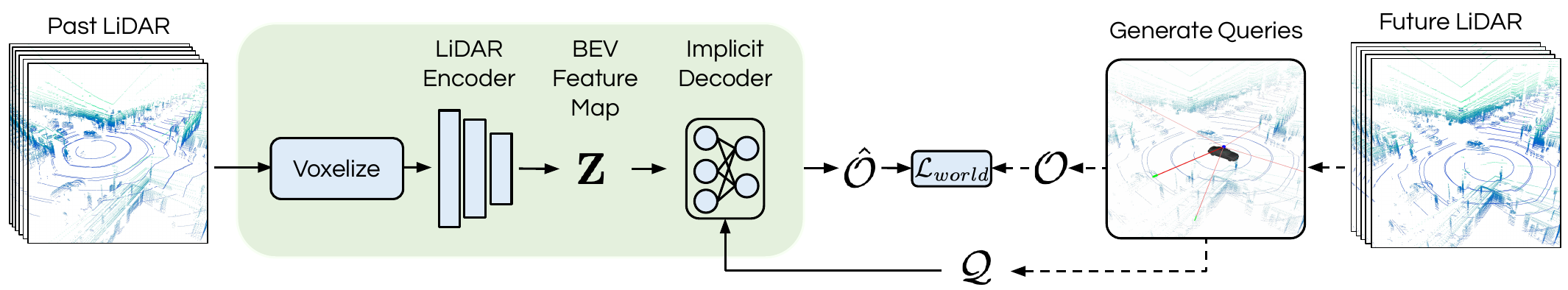}
    \vspace{-3mm}
    \caption{
        An overview of our method, \ourmodel{}.
        The past LiDAR is voxelized and encoded into a BEV feature map  
        which is used by an implicit occupancy decoder to predict occupancy $\hat{\mathcal{O}}$
        at query points $\mathcal{Q}$. 
        During training the query points and occupancy pseudo-labels are generated from future LiDAR data. 
        At inference, the model can be queried at any $(x, y, z, t)$ point.
        Refer to \cref{fig:objective} for details on the query generation process. 
    }
    \vspace{-1em}
    \label{fig:arch}
\end{figure*}

\ourmodel{} models 4D occupancy as a \textit{continuous field}, which has many advantages. First, continuous temporal resolution allows the model to properly handle the continuous nature of the rays' emission time during training, and be queried at any time of interest during inference.
Second, a very fine-grained spatial resolution is useful to accurately model complex shapes --- especially for thin structures --- without introducing quantization errors. Third, a fine spatial resolution allows us to make the best use of the self-supervision for positive occupancy for a small range interval $\delta$, without exceeding memory limits when dealing with very fine quantization in voxel-based methods. 

\paragraph{Architecture:}

Inspired by supervised implicit 3D occupancy~\cite{agro2023implicit}, we design a \textit{4D implicit occupancy forecasting architecture} that can predict occupancy at any arbitrary continuous query point $\mathbf{q} = (x, y, z, t)$ in 3D space and future time.
We refer the reader to \cref{fig:arch}. 

We first voxelize a stack of $H$ past LiDAR sweeps in BEV \cite{yang2018pixor} to represent the history, and then encode it through a ResNet \cite{he2016deep} backbone to generate a 2D BEV feature map, $\mathbf{Z}$.
This is preferable over a 3D feature volume since more representation and computing power can be allocated to the difficult task of learning dynamics in BEV, since objects primarily move along the relatively flat ground.

The BEV feature map $\mathbf{Z}$ is then used by an \textit{implicit occupancy decoder}, where
each 4D query point $\mathbf{q}=(x, y, z, t)$
is embedded with a positional embedding and outputs an occupancy logit by exploiting deformable attention \cite{zhu2020deformable} to relevant parts of the feature map.
This implicit decoder is very lightweight, so it can rapidly produce occupancy at many query points in parallel, allowing for practical training and efficiency for downstream tasks that require many query points.
More details are provided in the supplementary.

We note that the future sensor extrinsics are not an input to the model:
while they are used to accurately supervise the occupancy field as described in \cref{sec:method-selfsupervision}, it is desirable that the model is uncertain about the raycasting origin. This encourages the model to learn the extent of objects such that when viewed from any perspective, the future occupancy can explain the future LiDAR observations accurately. 

\paragraph{Training:}
A training example consists of LiDAR data from $t \in [t' - T_h, t' + T_f]$, where the history data $t \in [t' - T_h, t']$ is the model input and the future data $t \in [t', t' + T_f]$ is to be used as supervision by transforming it into 4D occupancy pseudo-labels as introduced in \cref{sec:method-selfsupervision}. Here, $T_p$ is the past time horizon and $T_f$ is the future time horizon.

To generate the query points used during training, we sample $N^+$ positive (occupied) points randomly along all positive rays $\mathcal{R}_{ij}^+$ that belong to $t_{ij} \in [t', t' + T_f]$ to form a set of points $\mathcal{Q}^+$ in occupied regions.
Similarly,
we sample $N_-$ negative (unoccupied) points uniformly along all current and future negative rays $\mathcal{R}_{ij}^-$ to form $\mathcal{Q}^-$.
As shown in \cref{fig:objective}, for most LiDAR points the $\mathcal{R}_{ij}^+$ positive ray segment is much shorter than
the negative ($\mathcal{R}_{ij}^-$), but setting $N^+ = N^-$ allows us to balance the positive and negative supervision during training. 
This generates a set of query points $\mathcal{Q} = \mathcal{Q}^+ \cup \mathcal{Q}^- = \{\mathbf{q}\}$ of cardinality $|\mathcal{Q}| = N^+ + N^-$.

We train our world model, with parameters $\theta$, using a simple binary cross-entropy loss summed across all query points. Let $f_\theta(\mathbf{q})$ denote \ourmodel{}'s occupancy probability output at a query point $\mathbf{q}$. Our self-supervised loss is:
\begin{align}
    \mathcal{L}_\text{world} = - \frac{1}{|\mathcal{Q}|}\left(\sum_{\mathbf{q} \in \mathcal{Q}^-} \log(1 - f_\theta(\mathbf{q})) + \sum_{\mathbf{q} \in \mathcal{Q}^+} \log(f_\theta(\mathbf{q})) \right).
\end{align}

\section{Transferring \ourmodel{} to downstream tasks} \label{sec:downstream}

To showcase the representational power and transferability of \ourmodel{}, we propose simple ways to transfer it to effectively perform the downstream tasks of point cloud forecasting and BEV semantic occupancy prediction.

\input{figures/main_qualitative.tex}
\subsection{Point Cloud Forecasting} \label{sec:downstream-point-cloud}

Point cloud forecasting has emerged as a natural task to evaluate the 4D understanding of self-driving world models \cite{weng2021inverting,weng2022s2net,mersch2022self,khurana2023point}.
The task consists of predicting future LiDAR point clouds given
past LiDAR point clouds and known sensor intrinsics and extrinsics into the future. 
Given \textit{query ray}s (sensor origin, direction, and timestamp) from the future LiDAR returns, the goal is to predict the depth $d$ the rays will travel before hitting any surface.

We can simply transfer \ourmodel{} to perform this task by 
learning a lightweight neural network $g_\gamma(\cdot)$ that acts as a renderer and predicts ray depth $d$ from a vector of UnO-estimated occupancy values along a query ray.
Specifically, to predict depth given a query ray, we first generate $N_r = 2000$ query points at a fixed interval $\epsilon_r = \SI{0.1}{m}$ along the ray, with the
time of each query point equal to the timestamp of the query ray.
\ourmodel{} is queried at each of these points, to generate $N_r$ occupancy predictions, which we concatenate into a vector
$\hat{\mathbf{o}} = [\hat{o}_1, \dots, \hat{o}_{N_r}]$. 
During training, we sample a \textit{batch} composed of a random subset of rays from the future point clouds in order to avoid going out of memory.
For the objective, we use a simple $\ell_1$ loss against the ground truth (GT) depth:
\begin{align}
    \mathcal{L}_{render} = \sum_{(d_{GT}, \hat{\mathbf{o}}) \in \text{ batch}} || d_{GT} - g_\gamma(\hat{\mathbf{o}}) ||_1.
\end{align}
More details on the network $g_\gamma(\cdot)$ are provided in the supplementary. 
Note that we freeze \ourmodel{} and train the renderer separately during this transfer, so the gradients do not affect the world model.

\subsection{BEV Semantic Occupancy Forecasting}  \label{sec:method-finetune}

BEV semantic occupancy forecasting, which consists of predicting the occupancy probability for a set of predefined classes (e.g., vehicles, cyclists, pedestrians) for multiple time steps into the future, has become a popular task in recent years \cite{sadat2020perceive,hu2021fiery,casas2021mp3,mahjourian2022occupancy,agro2023implicit,exquisite}.
The main reasons are that planning for self-driving is usually performed in BEV as the vast majority of objects of interest move along this plane, and that vulnerable road users such as pedestrians are more difficult to perceive and forecast due to irregular shapes and non-rigid motion, challenging the assumptions of traditional object detection and trajectory prediction.

In this section, we describe an efficient and simple fine-tuning method to transfer \ourmodel{} to this task by leveraging supervision from semantic object annotations.
To be able to query BEV points $\mathbf{q}_{BEV} = (x, y, t)$, we take a pretrained \ourmodel{} and replace the input $z$ with a fixed learnable parameter $\psi$.
Then, the original parameters from \ourmodel{} ($\theta$) along with the new parameter ($\psi$) are jointly fine-tuned in a second training stage on a small set of labeled data.
Following ImplicitO \cite{agro2023implicit}, BEV query points are randomly sampled continuously in space and future time, and are supervised with a binary cross-entropy loss per class indicating whether the BEV point lies within an object of that category or not.

\section{Experiments}

In this section, we evaluate the performance of \ourmodel{} across multiple tasks and datasets.
Point cloud forecasting (\cref{sec:point-cloud-forecasting-exp}) provides an increasingly popular and active benchmark where we can compare against prior self-driving world models.
BEV semantic occupancy prediction is another very interesting task because of its direct applicability to motion planning for self-driving \cite{sadat2020perceive,casas2021mp3,hu2021fiery,exquisite}
(\cref{sec:finetuning-exp}).
Finally, we also evaluate the quality of the geometric 4D occupancy the world model predicts (before any transfer) compared to other occupancy-based world models, with an emphasis on how well different methods can predict occupancy for object classes relevant to self-driving (\cref{sec:occupancy-recall-classes}).

\begin{table}[]
    \vspace{-10pt}
    \setlength\tabcolsep{2pt} %
    \centering
    \footnotesize
    \begin{tabular}{@{}l | lllll@{}}
    \toprule 
    \multicolumn{1}{c}{}                                                           &  \multicolumn{1}{c}{}               & NFCD (\SI{}{\meter\squared}) $\downarrow$ & CD (\SI{}{\meter\squared}) $\downarrow$ & AbsRel (\SI{}{\percent}) $\downarrow$ & L1 (\SI{}{\meter}) $\downarrow$ \\ \midrule
    \parbox[t]{2mm}{\multirow{3}{*}{\rotatebox[origin=c]{90}{\textbf{AV2}}}}       & RayTracing \cite{khurana2023point}  & 2.50                                      & 11.59                                   & 25.24                                 & 3.72                                    \\
                                                                                   & \CMU \cite{khurana2023point}        & 2.20                                      & 69.81                                   & 14.62                                 & 2.25                                    \\
                                                                                   & \ourmodel{}                         & \textbf{0.71}                             & \textbf{7.02}                           & \textbf{8.88}                         & \textbf{1.86}                           \\ \bottomrule
                                                                                   \toprule
    \parbox[t]{2mm}{\multirow{5}{*}{\rotatebox[origin=c]{90}{\textbf{NuScenes}}}}  & SPFNet \cite{weng2021inverting}     & 2.50                                      & 4.14                                   & 32.74                                 & 5.11                                    \\
                                                                                   & S2Net \cite{weng2022s2net}          & 2.06                                      & 3.47                                   & 30.15                                 & 4.78                                    \\
                                                                                   & RayTracing \cite{khurana2023point}  & 1.66                                      & 3.59                                   & 26.86                                 & 2.44                                    \\
                                                                                   & \CMU \cite{khurana2023point}        & 1.40                                      & 4.31                                   & 13.48                                 & 1.71                                    \\
                                                                                   & \ourmodel{}                         & \textbf{0.89}                             & \textbf{1.80}                          & \textbf{8.78}                         & \textbf{1.15}                           \\ \bottomrule
                                                                                   \toprule
    \parbox[t]{2mm}{\multirow{3}{*}{\rotatebox[origin=c]{90}{\textbf{KITTI}}}}     & ST3DCNN \cite{mersch2022self}       & 4.19                                      & 4.83                                   & 28.58                                  & 3.25                                    \\
                                                                                   & \CMU \cite{khurana2023point}        & 0.96                                      & 1.50                                   & 12.23                                  & 1.45                                    \\
                                                                                   & \ourmodel{}                         & \textbf{0.72}                             & \textbf{0.90}                          & \textbf{9.13}                         & \textbf{1.09}                           \\ \bottomrule
    \end{tabular}
    \vspace{-2mm}
    \caption{Point cloud prediction results on Argoverse 2 LiDAR, NuScenes, and KITTI.}
    \label{tab:point-cloud}
    \vspace{-3mm}
\end{table}

\paragraph{Implementation details:}
We conduct our experiments on Argoverse 2 \cite{wilson2023argoverse}, nuScenes \cite{caesar2020nuscenes}, and KITTI Odometry \cite{behley2019semantickitti, geiger2013vision}, three popular large-scale datasets for autonomous driving. 
On KITTI and Argoverse2, \ourmodel{} and the baselines receive $H = 5$ 
past frames at an interval of $\SI{0.6}{s}$.
For nuScenes, the input is $H = 6$ at an interval of $\SI{0.5}{s}$.
These settings follow prior work in point cloud forecasting \cite{khurana2023point}.
For all datasets, we use a learning rate of $8.0 \times 10^{-4}$ with a 1000 iteration warmup
from a learning rate of $8.0 \times 10^{-5}$ and a cosine learning rate schedule.
We train for a total of 50,000 iterations with a batch size of 16
on the training split of each dataset with the AdamW optimizer.
On each batch sample, we train \ourmodel{} on $N^+ = N^- = 900,000$ positive and negative query points.
The implicit occupancy decoder is very lightweight, so it can run efficiently on many queries in parallel:
\ourmodel{} has roughly 17.4M parameters, only 0.06M of which are in the implicit occupancy decoder.
Since some baselines are task-specific we introduce them in the corresponding sections. 
More details for model hyperparameters, training process for each dataset, and baselines can be found in the supplementary.

\subsection{Point Cloud Forecasting} \label{sec:point-cloud-forecasting-exp}
Following \CMU{} \cite{khurana2023point}, we utilize Chamfer distance (CD), Near Field Chamfer Distance (NFCD), depth L1 error (L1), depth relative L1 error (AbsRel) as our metrics.
NFCD computes CD on points inside the region of interest (ROI), which is defined to be $\left[-70, 70\right]$ \SI{}{m} in both $x$-axis and $y$-axis, as well as 
$\left[-4.5, 4.5\right]$ \SI{}{m} in $z$-axis around the ego vehicle.
AbsRel is just the L1 error divided by the ground-truth depth.
Models are evaluated on \SI{3}{s} long point cloud forecasts.
For the Argoverse 2 LiDAR and KITTI datasets, the target point clouds are evaluated
at $\{\SI{0.6}{s}, \SI{1.2}{s}, \dots, \SI{3.0}{s}\}$, and for 
nuScenes they are at $\{\SI{0.5}{s}, \SI{1.0}{s}, \dots, \SI{3.0}{s}\}$.
While these parameters are chosen to match prior evaluation protocols \cite{khurana2023point},
\ourmodel{} can produce occupancy and point cloud predictions at any continuous point
in space and future time without re-training thanks to its implicit architecture.
This is important for downstream tasks such as motion planning which may require occupancy at arbitrary times.

\paragraph{Comparison against state-of-the-art:}
\cref{tab:point-cloud} shows quantitative comparisons of \ourmodel{} and the state-of-the-art unsupervised point cloud forecasting methods. 
\ourmodel{}, which is transferred to this task as explained in \cref{sec:downstream-point-cloud}, demonstrates significant improvements in all metrics and
all datasets.
We observe that \ourmodel{} captures dynamic objects better than the baselines, which we investigate further in \cref{sec:occupancy-recall-classes}. This explains
the largest relative improvement on Argoverse 2, where there are more dynamics objects \cite{wilson2023argoverse}. 
Additionally, we submitted UnO to the public Argoverse 2 leaderboard\footnote{\href{https://eval.ai/web/challenges/challenge-page/1977/leaderboard/4662}{https://eval.ai/web/challenges/challenge-page/1977/leaderboard/4662}}
for point cloud forecasting, where we achieved first place.
\cref{fig:main-qualitative} visualizes the point cloud predictions of our model,
where we see it accurately forecasts moving vehicles with diverse future behaviors,
and it captures small objects. 
See the supplementary for a visual comparison of our model to baselines and ablations, and more details about the public leaderboard.

\begin{table}[]
    \vspace{-10pt}
    \setlength\tabcolsep{2pt} %
    \centering
    \scriptsize
    \begin{tabular}{llllllll}
    \toprule 
                                                                             & \multicolumn{4}{c}{\textbf{Point Cloud Forecasting}}                                                                                                                                                                                                                                                                                          & \multicolumn{2}{c}{\begin{tabular}[c]{@{}l@{}}\textbf{BEV Semantic}\\\textbf{Occupancy}\end{tabular}}                                                              \\
                                                                             \cmidrule(l{5pt}r{5pt}){2-5} \cmidrule(l{5pt}){6-7}
    \begin{tabular}[c]{@{}l@{}}Pre-training\\Procedure\end{tabular}          & \begin{tabular}[c]{@{}l@{}}NFCD\\ (\SI{}{\meter\squared}) $\downarrow$\end{tabular} & \begin{tabular}[c]{@{}l@{}}CD\\ (\SI{}{\meter\squared}) $\downarrow$\end{tabular} & \begin{tabular}[c]{@{}l@{}}AbsRel\\ (\SI{}{\percent}) $\downarrow$\end{tabular} & \begin{tabular}[c]{@{}l@{}}L1\\ (\SI{}{\meter\squared}) $\downarrow$\end{tabular} & \begin{tabular}[c]{@{}l@{}}mAP\\$\uparrow$\end{tabular}      & \begin{tabular}[c]{@{}l@{}}Soft-IoU\\$\uparrow$\end{tabular}  \\ \midrule
    \depthrendering{}                                                        & 1.79                                                                                & 14.1                                                                              & 17.0                                                                            & 2.80                                                                              & 20.7                                                         & 8.6                                                           \\ 
    \freespacerendering{}                                                    & 1.16                                                                                & 10.0                                                                              & 13.0                                                                            & 2.28                                                                              & 39.6                                                         & 14.6                                                           \\  %
    \unbalancedourmodel{}                                                    & 0.84                                                                                & 8.90                                                                               & 11.8                                                                            & 2.04                                                                              & 43.6                                                         & 17.0                                                          \\  %
    \ourmodel{}                                                              & \textbf{0.83}                                                                       & \textbf{8.10}                                                                     & \textbf{10.1}                                                                  & \textbf{2.03}                                                                     & \textbf{52.3}                                                & \textbf{22.3}                                                 \\ \bottomrule %
    \end{tabular}
    \vspace{-2mm}
    \caption{Ablating the effect of occupancy pre-training procedures on the downstream tasks of point cloud forecasting and BEV semantic occupancy forecasting in Argoverse 2 Sensor dataset.}
    \label{tab:pre-training-ablation}
    \vspace{-3mm}
\end{table}

\paragraph{Effect of the pre-training method:}  \label{sec:point_cloud_pretraining}
We ablate multiple pre-training procedures to better understand where our improvements over state-of-the-art come from,
keeping the architecture constant (as described in \cref{sec:method-UnO}).
The \freespacerendering{} objective used in \cite{khurana2022differentiable}
learns to forecast occupancy by leveraging visibility maps of future point clouds.
Concretely, the model computes free-space as the complement
of the cumulative maximum occupancy along each LiDAR ray.
The free-space predictions are trained via cross entropy to match the visibility map.
The \depthrendering{} objective is the differentiable depth rendering loss
used in \cite{khurana2023point}.
This objective is similar to the one used in NeRF~\cite{mildenhall2021nerf}, 
but for estimating the expected ray depth instead of the color, based on a sequence of occupancy predictions along the ray at a fixed depth interval of $\epsilon_r$.
The \unbalancedourmodel{} objective is the same as the
\ourmodel{} objective described in \cref{sec:method-UnO}, but the 
query points are sampled uniformly along the ray $\{\mathcal{R}_{ij}^-,\mathcal{R}_{ij}^+\}$,
instead of an equal number of query points with positive and negative labels.

While \depthrendering{} directly estimates ray depth, \ourmodel{} and \freespacerendering{} do not.
Thus, we train the learned renderer described in \cref{sec:downstream} for each of these occupancy forecasting models.
\cref{tab:pre-training-ablation} presents the results 
of this comparison, evaluated on the Argoverse 2 Sensor dataset. 
We notice that \ourmodel{} greatly outperforms the baselines on all metrics.
Although the gains of balancing the loss are small in point cloud forecasting metrics, we can see the importance of it for BEV semantic occupancy, which we discuss further in \cref{sec:finetuning-exp}.

\subsection{Semantic 3D Occupancy Forecasting (BEV)} \label{sec:finetuning-exp}

\begin{figure}
    \centering
    \vspace{-5pt}
    \begin{tikzpicture}
    \definecolor{clr1}{HTML}{ef9b20}
    \definecolor{clr2}{HTML}{bdcf32}
    \definecolor{clr3}{RGB}{0,160,236}
    \definecolor{clr4}{HTML}{E88EED}
    \begin{axis}[
        xlabel=Number of labeled training frames,
        ylabel=mAP,
        xmode=log,
        log ticks with fixed point,
        legend pos=south east,
        legend style={nodes={scale=0.8, transform shape}, draw=none}, 
        width=\linewidth,
        height=6cm,
        axis lines=left,
        xmin=0.5,
        xmax=100000,
        ymin=-5,
        ymax=80,
        grid=both,
        grid style={line width=.1pt, draw=gray!30},
    ]

    \addplot[color=clr1, mark=*, mark size=1.5pt, thick] coordinates {
        (1, 1.8) %
        (10, 2.1) %
        (100, 10.8) %
        (1000, 39.8)  %
        (10000, 65.5) %
        (67000, 69.0) %
    };
    \addlegendentry{MP3}

    \addplot[color=clr2, mark=*, mark size=1.5pt, thick] coordinates {
        (1, 2.3)
        (10, 2.8) %
        (100, 12.7) %
        (1000, 41.3) %
        (10000, 67.9) %
        (67000, 70.9) %
    };
    \addlegendentry{ImplicitO}

    \addplot[color=clr3, mark=*, mark size=1.5pt, thick] coordinates {
        (1, 22.1)
        (10, 44.5) %
        (100, 54.2) %
        (1000, 63.2) %
        (10000, 73.9) %
        (67000, 74.6) %
    };
    \addlegendentry{\ourmodel{}}

    \end{axis}
    \end{tikzpicture}
    \vspace{-7mm}
    \caption{
        BEV semantic occupancy results. Fine-tuning \ourmodel{} vs. SOTA supervised methods across different scales of supervision. 
    }
    \label{fig:finetuning}
    \vspace{-2mm}
\end{figure}
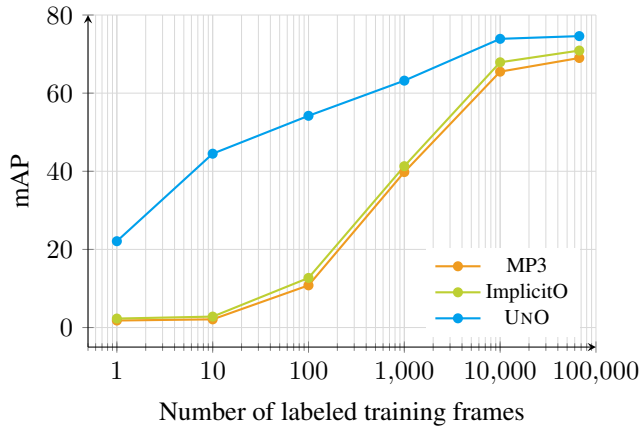

In this section we show that pre-training with unsupervised occupancy can be useful for the downstream task of 3D semantic occupancy forecasting in BEV (2D + time), especially when the supervision available is limited.
We follow the evaluation protocol of \cite{agro2023implicit},
which evaluates occupancy forecasts on annotated vehicles from the
Argoverse 2 Sensor dataset. 
Unlike Argoverse 2 LiDAR, this dataset has object labels annotations.
For BEV semantic occupancy supervision and evaluation, we consider a ROI
of \SI{80}{m} by \SI{80}{m} centered on the ego vehicle, and a future time
horizon of \SI{3}{s}.
The \ourmodel{} model that serves as pre-trained weights for this experiment is the same as the one used in \cref{sec:point-cloud-forecasting-exp}.
During supervised fine-tuning (explained in \cref{sec:method-finetune}), the query-points are sampled uniformly at
random from space and future time, and during evaluation the
query points are sampled on a uniform grid of spatial resolution \SI{0.4}{m}
at times $\{\SI{0.0}{s}, \SI{0.5}{s}, \dots, \SI{3.0}{s}\}$.
We follow \cite{mahjourian2022occupancy,agro2023implicit} in the use of
mean Average Precision (mAP) and Soft-IoU as metrics to evaluate BEV
semantic occupancy prediction.

\paragraph{Benchmark results:}
\cref{fig:finetuning} compares 
\ourmodel{} to SOTA BEV semantic occupancy prediction models MP3 \cite{casas2021mp3} and 
ImplicitO \cite{agro2023implicit} across varying amounts of labeled
data, spanning from just 1 labeled frame to the entire Argoverse 2 Sensor
\texttt{train} split. We train all models until their validation 
loss plateaus. 
When finetuning \ourmodel{}, it outperforms the baselines
at all levels of supervision, including when all labels are available.
In the few-shot learning setting (1-10 frames), the results are particularly impressive, showing the representational power and transferability of \ourmodel{}.
Even with a single frame of semantic supervision, our model can learn to separate vehicles from other classes and background to a certain degree.
We also call out that with roughly an order of magnitude less labeled data,
\ourmodel{} outperforms MP3 and ImplicitO trained with all the available training data. 
This highlights the efficacy of our pre-training procedure for understanding the geometry, dynamics and semantics of the scene, and the applicability of this understanding to this important task in the self-driving stack.

\input{figures/bev_semantic_occ.tex}

\begin{figure*}
    \centering
    \vspace*{-10pt}
    \begin{tikzpicture}
        \definecolor{clr1}{HTML}{ef9b20}
        \definecolor{clr2}{HTML}{00c660}
        \definecolor{clr3}{RGB}{0,160,236}
        \definecolor{clr4}{HTML}{E88EED}
        \definecolor{clr5}{HTML}{f5e45f}
        \begin{axis}[
            ybar,
            bar width=0.07cm,
            enlarge x limits=0.05,
            ylabel={\footnotesize{Unsupervised Occupancy Recall}},
            width=\textwidth,
            height=5cm,
            xtick=data,
            ybar=1pt, %
            xmin=3,
            xmax=28,
            ymin=0,
            ymax=100,
            axis lines=left,
            xticklabels={
                \scriptsize Wheelchair$^{*\dagger}$, %
                \scriptsize Stroller$^{*\dagger}$, %
                \scriptsize School Bus$^*$, %
                \scriptsize Dog$^{*\dagger}$, %
                \scriptsize Bicyclist, %
                \scriptsize Wheeled Rider$^*$, %
                \scriptsize Truck Cab, %
                \scriptsize Truck, %
                \scriptsize Motorcycle, %
                \scriptsize Motorcyclist$^*$, %
                \scriptsize Large Vehicle, %
                \scriptsize Bus, %
                \scriptsize Vehicular Trailer, %
                \scriptsize Box Truck, %
                \scriptsize Sign, %
                \scriptsize Wheeled Device, %
                \scriptsize Stop Sign, %
                \scriptsize Bicycle, %
                \scriptsize Construction Barrel$^\dagger$, %
                \scriptsize Construction Cone$^\dagger$, %
                \scriptsize Bollard$^\dagger$, %
                \scriptsize Pedestrian, %
                \scriptsize Regular Vehicle, %
                \scriptsize All, %
            },
            x tick label style={rotate=30,anchor=east},
            legend style={at={(0.05, 0.90)},anchor=south west, legend columns=-1, column sep=3pt, draw=none},
            ]

            \addplot[fill=clr4, color=clr4] coordinates {
                (4,  22.7)
                (5,  11.0)
                (6,  6.2)
                (7,  2.4)
                (8,  5.8)
                (9,  12.0)
                (10, 13.1)
                (11, 5.7)
                (12, 11.7)
                (13, 6.1)
                (14, 9.6)
                (15, 6.2)
                (16, 12.6)
                (17, 9.0)
                (18, 20.8)
                (19, 8.5)
                (20, 10.8)
                (21, 10.1)
                (22, 28.1)
                (23, 13.1)
                (24, 28.3)
                (25, 12.5)
                (26, 10.9)
                (27, 10.3)
            };
            \addlegendentry{\footnotesize{\CMU{} \cite{khurana2023point}}}

            \addplot[fill=clr2, color=clr2] coordinates {
                (4,  22.5)
                (5,  5.9)
                (6,  44.4)
                (7,  0.0)
                (8,  0.0)
                (9,  52.8)
                (10, 0.0)
                (11, 5.8)
                (12, 4.8)
                (13, 2.0)
                (14, 46.0)
                (15, 22.7)
                (16, 12.8)
                (17, 20.8)
                (18, 43.9)
                (19, 5.8)
                (20, 15.1)
                (21, 28.3)
                (22, 38.5)
                (23, 0.0)
                (24, 32.1)
                (25, 23.5)
                (26, 33.8)
                (27, 32.1)
            };
            \addlegendentry{\footnotesize{\depthrendering{}}}

            \addplot[fill=clr5, color=clr5] coordinates {
                (4,  21.7)
                (5,  5.9)
                (6,  63.1)
                (7,  2.7)
                (8,  2.9)
                (9,  58.3)
                (10, 59.0)
                (11, 66.3)
                (12, 17.4)
                (13, 17.7)
                (14, 58.5)
                (15, 57.4)
                (16, 61.5)
                (17, 50.7)
                (18, 44.4)
                (19, 9.3)
                (20, 20.9)
                (21, 38.7)
                (22, 41.4)
                (23, 3.7)
                (24, 41.2)
                (25, 28.9)
                (26, 47.6)
                (27, 47.8)
            };
            \addlegendentry{\footnotesize{\freespacerendering{}}}

            \addplot[fill=clr1, color=clr1] coordinates {
                (4,  43.3)
                (5,  11.8)
                (6,  57.3)
                (7,  29.7)
                (8,  3.6)
                (9,  83.3)
                (10, 14.7)
                (11, 51.6)
                (12, 13.5)
                (13, 10.6)
                (14, 56.5)
                (15, 51.5)
                (16, 30.0)
                (17, 34.8)
                (18, 19.9)
                (19, 15.8)
                (20, 28.5)
                (21, 35.9)
                (22, 55.6)
                (23, 66.6)
                (24, 62.8)
                (25, 35.7)
                (26, 49.8)
                (27, 48.4)
            };
            \addlegendentry{\footnotesize{\unbalancedourmodel{}}}

            \addplot[fill=clr3, color=clr3] coordinates {
                (4,  65.8)
                (5,  52.9)
                (6,  80.7)
                (7,  41.9)
                (8,  70.0)
                (9,  83.3)
                (10, 26.1)
                (11, 68.3)
                (12, 72.5)
                (13, 84.3)
                (14, 78.5)
                (15, 79.9)
                (16, 60.4)
                (17, 75.6)
                (18, 60.2)
                (19, 34.6)
                (20, 48.8)
                (21, 53.0)
                (22, 69.8)
                (23, 70.4)
                (24, 74.9)
                (25, 64.2)
                (26, 66.7)
                (27, 67.0)
            };
            \addlegendentry{\footnotesize{\ourmodel{}}}

        \end{axis}
    \end{tikzpicture}
    \vspace{-4mm}
    \caption{
        Unsupervised occupancy recall comparison on the Argoverse 2 Sensor dataset, averaged across the prediction horizon. Recall is computed at a precision of 0.7.
        $^*$ denotes the rarest $25\%$ of classes, and $^\dagger$ denotes the smallest (by bounding box volume) $25\%$ of classes.
    }
    \label{fig:classes-bar-plot-vs-all}
    \vspace{-2mm}
\end{figure*}
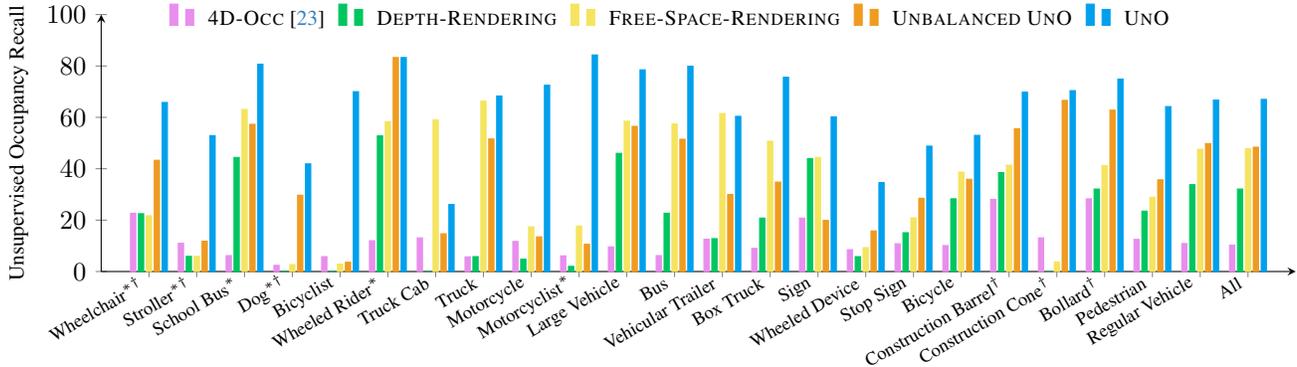

\paragraph{\ourmodel{} pre-training is key to understanding vehicles:}
For each pre-trained model in \cref{tab:pre-training-ablation},
we fine-tune it on 100 frames of labeled data from the Argoverse 2
Sensor dataset using the procedure described in
\cref{sec:downstream}, with the encoder frozen for faster convergence. 
The occupancy forecasting metrics are in
\cref{tab:pre-training-ablation}.
We observe that \ourmodel{} provides pre-trained weights that are best for fine-tuning for BEV semantic occupancy, outperforming other pre-training procedures by a large margin.
While \unbalancedourmodel{} achieves similar point cloud forecasting
performance to \ourmodel{}, it is significantly worse when transferred to BEV semantic occupancy forecasting.
The large ratio of free to occupied space means that the 
\unbalancedourmodel{} gets less supervision on difficult moving 
objects, which makes learning accurate dynamic occupancy and object extent more difficult
(see the supplementary for visualizations).
Point cloud forecasting metrics are largely dominated by the static background and occupancy extent does not
matter for the learned raycasting model, explaining the similarity in performance between
\ourmodel{} and \unbalancedourmodel{} in these metrics.
However, \ourmodel{} has learned a richer representation of moving objects and extent, which are both important
for BEV semantic occupancy prediction.

\paragraph{Qualitative results:} 
\cref{fig:bev-semantic-occ} illustrates BEV semantic occupancy predictions of fine-tuned \ourmodel{}
on all available labeled data of vehicles from the training split of Argoverse 2 Sensor.
\ourmodel{} perceives all vehicles in the scene, and accurately predicts their behavior into the future.
We notice that \ourmodel{} implicitly understands the map, although it is not used as input or supervision.
This is likely because our self-supervision of geometric occupancy forces the model to learn
the roadway in order to predict accurate future behavior.

\vspace{-1ex}
\subsection{Geometric 4D Occupancy Forecasting} \label{sec:occupancy-recall-classes}

\cref{sec:point-cloud-forecasting-exp} showed that \ourmodel{} brings significant advantages to point cloud forecasting.
However, point cloud forecasting metrics are dominated by background areas where most of the LiDAR points are. 
In practice, these areas are the least
relevant to downstream self-driving tasks like motion planning,
where accurately modelling foreground dynamic road agents is vital. 
While \cref{sec:finetuning-exp} already shed some light onto \ourmodel{}'s understanding of dynamic objects, in this section we seek to evaluate the unsupervised geometric occupancy predictions directly, without any transfer.

To evaluate the 4D occupancy quality on relevant objects, we compute the occupancy recall within the labeled 3D bounding boxes (each with a semantic class) provided by the Argoverse 2 Sensor \texttt{val} split.
To ensure fair comparison between models, we evaluate the recall of all models at the same 
precision level. For each model, we determine the occupancy confidence threshold that achieves a precision of 70\%. 
To compute precision, we consider all points inside the ground truth bounding boxes
(of any class) as positives, and the remainder of the space as negatives (excluding non-visible regions, which are ignored).
The visible regions at a given time step are determined using ray tracing
to compute a 3D visibility voxel grid from the LiDAR point cloud \cite{hu2020you}.
We evaluate points at a grid of size $\SI{80}{m}$ by 
$\SI{80}{m}$ centered on the ego in $x$ and $y$ and $[\SI{0.0}{m}, \SI{3.0}{m}]$
in $z$, using a spatial discretization of $\SI{0.2}{m}$ at future timesteps
$\{\SI{0.6}{s}, \SI{1.2}{s}, \dots, \SI{3.0}{s}\}$.

\paragraph{Class-wise 4D occupancy recall results:}
\cref{fig:classes-bar-plot-vs-all} shows the results of this experiment. 
We first note that \ourmodel{} succeeds in predicting occupancy for small and rare objects.
Furthermore, \ourmodel{} attains significantly higher recall than
state-of-the-art \cite{khurana2023point} and the other unsupervised occupancy ablations for almost all object classes, with impressive improvements on many vulnerable road users such as Stroller, Bicyclist, and Motorcycle/Motorcyclist.

Multiple aspects of \ourmodel{}'s geometric occupancy predictions contribute to the significant improvements over the prior art.
We observe an improved forecasting of
the behavior of dynamic objects. \ourmodel{}'s occupancy
captures the multi-modal behavior of pedestrians (who may either cross the road
or continue along the sidewalk), vehicles (which may lane change or change speeds),
and even bicyclists (who move inside and outside of traffic). See \cref{fig:main-qualitative} and the supplementary
for some examples. On the other hand, \CMU{}, \depthrendering{} and \freespacerendering{} struggle to accurately
model dynamic objects.
\unbalancedourmodel{} captures dynamic objects better than the other baselines, but not as well as \ourmodel{}; which makes sense as its supervision has less weight on moving foreground objects.
We note that \depthrendering{} uses the same objective function as
\CMU{} but has better recall in most categories.
We attribute this to \depthrendering{}'s implicit architecture (same as \ourmodel{}),
which helps mitigate discretization error and improve effective receptive field.
Moreover, \ourmodel{} better captures the extent of traffic participants,
going beyond their visible parts (see \cref{fig:main-qualitative}).
The other baselines, particularly \CMU{} and \freespacerendering{}, only capture the visible surfaces in the LiDAR point clouds, as shown in the supplementary.

\vspace{-1ex}
\section{Conclusion}
\vspace{-1ex}
In this paper, we propose \ourmodel{}, a powerful unsupervised occupancy world model that forecasts a 4D geometric occupancy field from past LiDAR data.
To tackle this problem, we leverage the occupancy implied by future point clouds as supervision to train an implicit architecture that can be queried at any continuous $(x, y, z, t)$ point.
Not only does \ourmodel{} achieve an impressive understanding of the geometry, dynamics and semantics of the world from unlabeled data, but it can also be effectively and easily transferred to perform downstream tasks.
To demonstrate this ability, we show that \ourmodel{} outperforms the state-of-the-art on the tasks of point cloud forecasting and supervised BEV semantic occupancy prediction.
We hope that \ourmodel{} and future work in unsupervised world models will unlock greater safety for self-driving, notably for vulnerable and rare road users.

{\small
\bibliographystyle{ieee_fullname}
\bibliography{egbib}
}

\clearpage

\onecolumn
\section*{Appendix}
\appendix

\setlist[itemize]{itemsep=3pt, leftmargin=20pt, topsep=8pt}
\setlist[enumerate]{itemsep=3pt, leftmargin=20pt, topsep=8pt}

In this appendix, we describe implementation details relevant to \ourmodel{} including
architecture and training, implementation details for any baselines and ablations,
and further details on metrics and evaluation procedures.
Finally, we showcase several additional results:
\begin{itemize}
    \item Visualizations comparing the 4D occupancy and point cloud forecasts of \ourmodel{} against various
    baselines including the SOTA \CMU{}.
    \item Visualizations of \ourmodel{}'s occupancy in the context of the camera data provided by Argoverse 2.
    \item Comparing the 4D occupancy recall of \ourmodel{}, \ourmodel{} fine-tuned
    with bounding box data, and \ourmodel{} trained with bounding box data without pre-training for 4D occupancy.
\end{itemize}

\section{\ourmodel{} Implementation Details} \label{sec:implementation-details}

\subsection{LiDAR Encoder}

\begin{figure*}[t]
    \centering
    \includegraphics[width=\textwidth]{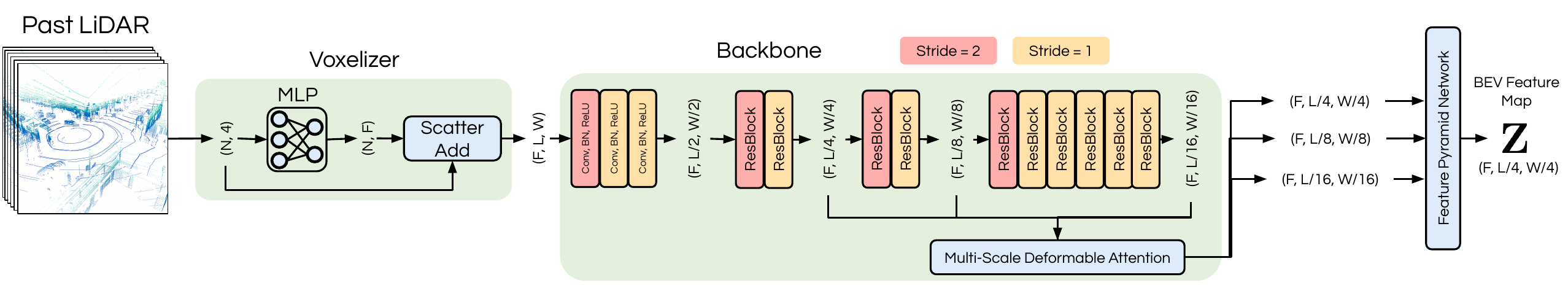}
    \caption{
        The architecture of the LiDAR Encoder. The batch dimension is omitted from the tensor shapes.
    }
    \label{fig:encoder}
\end{figure*}

The LiDAR encoder takes $H$ past LiDAR sweeps to produce a Birds Eye View (BEV) feature map $\mathbf{Z}$ used by the
implicit decoder of \ourmodel{}. 
Below we describe the architecture of the LiDAR encoder, illustrated in \cref{fig:encoder}, where we omit the batch dimension
for simplicity.

\paragraph{Voxelization:}

Each LiDAR sweep is a set of points with four features $(x, y, z, t)$. 
A small MLP is used encode these points as feature vectors of size $F = 128$.
The LiDAR points are placed in a 2D BEV grid of shape $(L, W)$ based on their $(x, y)$ coordinates. We use an ROI of
$[-100, 150]\;\text{m}$ on the $x$ dimension, $[-100, 100]\;\text{m}$ on the $y$ dimension with a voxel
size of $\SI{0.15625}{m}$ in $x, y$, resulting in $L = \frac{200}{0.15265} = 1280, W = \frac{250}{0.1526} = 1600$. 
A 2D feature map of shape $(F, L, W)$ is then generated from this BEV grid, where each grid cell has a feature vector
that is the sum of the features from all points in that grid cell.

\paragraph{Backbone:}

This 2D feature map is processed by three convolution layers interleaved with batch-normalization and ReLU layers,
the first convolutional layer having a stride of 2, resulting in a feature map of shape $(F, L/2, W/2)$.

Then, this feature map is processed by a series of ten residual layers, labelled \textit{ResBlock} in \cref{fig:encoder}, which each employ a sequence of dynamic convolution \cite{chen2020dynamic}, batch-normalization, ReLU, dynamic convolution, batch-normalization, squeeze-and-excitation \cite{hu2018squeeze}, and dropout operations. Each residual layer produces a feature map, and layers 0, 2, and 4 down-sample their output by a factor of 2 (using convolutions with stride 2).

We extract three multi-level feature maps from the output of layers 1, 3, and 9 with shapes $(F, L/4, W/4)$, $(F, L/8, W/8)$, and $(F, L/16, W/16)$
respectively,
across which information is fused with multiscale deformable attention \cite{zhu2020deformable} to
produce three feature maps with the same shapes as the input. %

\paragraph{Feature Pyramid Network}

These three multi-level feature maps are fused into a single feature map $\mathbf{Z}$ of shape $(F, L/4, W/4)$
by a lightweight Feature Pyramid Network \cite{lin2017feature}. $\mathbf{Z}$ is used by the implicit decoder,
described below, to predict 4D geometric occupancy. 

\subsection{Implicit Occupancy Decoder}

\begin{figure*}[t]
    \centering
    \includegraphics[width=\textwidth]{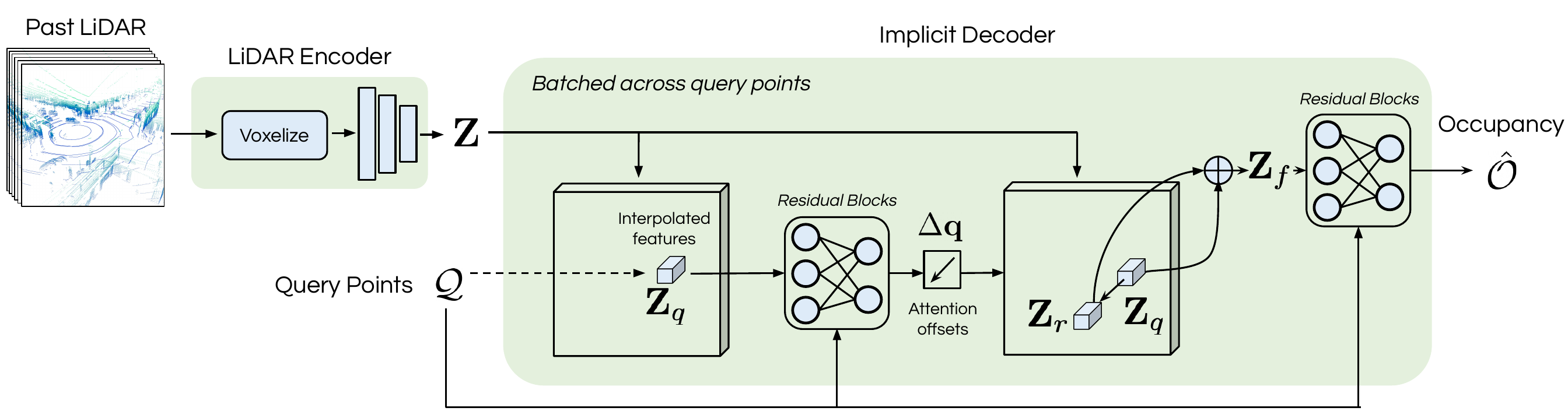}
    \caption{
        \ourmodel{}'s architecture zoomed in on the implicit decoder.
    }
    \label{fig:decoder}
\end{figure*}

The inputs to the implicit occupancy decoder are 
the feature map $\mathbf{Z} \in \mathbb{R}^{F \times \frac{L}{4} \times \frac{W}{4}}$ from the lidar encoder, and
a set of 4-dimensional query points $\mathcal{Q} \in \mathbb{R}^{|\mathcal{Q}| \times 4}$ with features $(x, y, z, t)$.
Our decoder follows closely that of ImplicitO \cite{agro2023implicit}, but we describe it below.
See \cref{fig:decoder} for an architecture diagram.
The decoder consists of three main parts: offset prediction, feature aggregation, and occupancy prediction.

\paragraph{Offset Prediction:}

The offset prediction module begins by interpolating $\mathbf{Z}$ at the $(x, y)$ locations specified by the
spatial components of the query points $\mathbf{q}_{x,y} \in \mathcal{Q}_{x,y}$ to produce interpolated feature vectors $\mathbf{z}_q$
that contain information about the scene around each query point. 
Next, linear projections of $\mathbf{Z}_q$ and $\mathbf{q}$ are added and passed through a
residual layer consisting of two fully connected linear layers and a residual connection.
We use a dimensionality of $16$ for these linear projections and
the hidden size of the linear layers in the residual layers.
Finally, a linear layer is used to produce an offset per query point
$\Delta \mathbf{q}$. These offsets are added to the query points to find the offset sample locations
$\mathbf{r} = \mathbf{q}_{x,y} + \Delta \mathbf{q}_{x,y}$, which is the input to the next module.

\paragraph{Feature Aggregation:}
The offset sample points $\mathbf{r}$ are meant to store the $(x', y')$ locations in $\mathbf{Z}$ that contain
important information for occupancy prediction at query point $(x, y, t)$.
As such, the feature aggregation module begins by interpolating $\mathbf{Z}$ at $\mathbf{r}$ to obtain a
feature vector $\mathbf{Z}_r$, for each query point.
This feature vector is concatenated with $\mathbf{z}_q$ to obtain the summary features for the query point,
denoted $\mathbf{Z}_f$, of dimensionality $2F = 256$. 

\paragraph{Occupancy Prediction:}

This module uses the aggregated feature vector $\mathbf{Z}_f$ and the query points $\mathbf{q}$
to predict occupancy logits, $\hat{\mathcal{O}}$, for all query points.
$\mathbf{Z}_f$ and $\mathbf{q} = (x, y, z, t)$ are passed through
three residual blocks with an architecture inspired by Convolutional Occupancy Networks \cite{peng2020convolutional}. 
The residual block takes as input a linear projection of $\mathbf{q} = (x, y, z, t)$.
Each block adds its input to a linear projection of $\mathbf{Z}_f$, and then passes this through a residual layer
using a hidden size $16$. The input to the subsequent block is the output of the previous block.

\subsection{Rendering Header} \label{sec:rendering-header}

\begin{figure*}[t]
    \centering
    \includegraphics[width=0.9\textwidth]{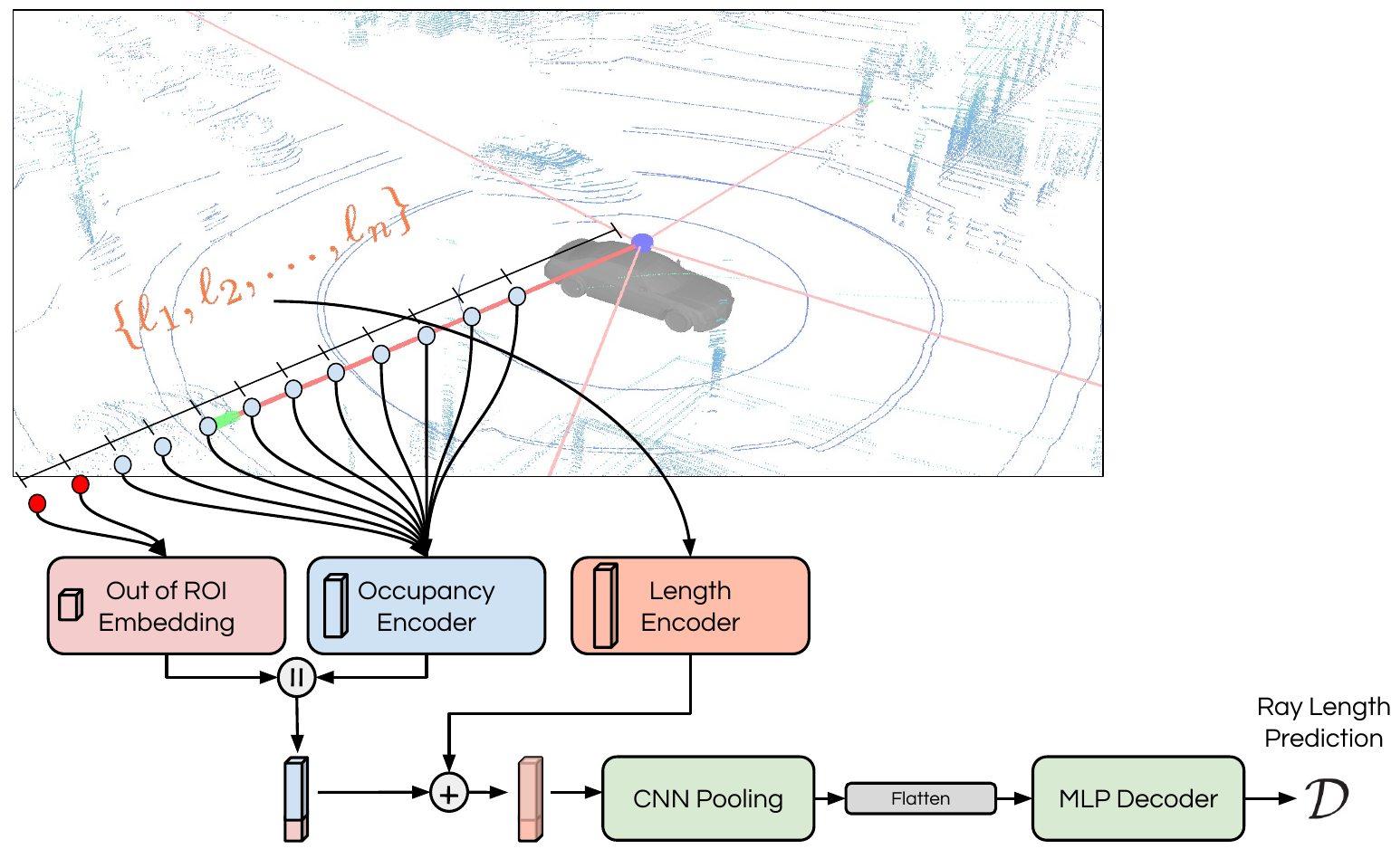}
    \caption{
        An overview of our the learned render header. The predicted occupancy at each point along the ray is combined with the distance of the point from the sensor, after which the features are reduced using a 1D CNN and then passed through an MLP to produce a single scalar prediction. 
        $||$ denotes lengthwise concatenation, and $+$ denotes addition.
    }
    \label{fig:render}
\end{figure*}

The rendering header is the small network that uses the occupancy predictions of \ourmodel{} along a
lidar ray and estimates the depth of the lidar point along the ray. 
Here we provide more details on the architecture and training of this header.
A diagram is in \cref{fig:render}.
Given a query lidar ray, we generate $N_r = 2000$ query points along the ray every $\epsilon_r = \SI{0.1}{m}$.
The query points inside \ourmodel{}'s ROI are fed to \ourmodel{}, which predicts occupancy logits
in a tensor of shape ${N_{r_{in}} \times 1}$ where
$N_{r_{in}}$ is the number of points inside the ROI. 
The occupancy logits are encoded by a linear layer (``Occupancy Encoder'' in \cref{fig:render}) to produce
a tensor of shape ${N_{r_{in}} \times 256}$.
We maintain a learned embedding of shape $N_r \times 256$, and for the $N_{r_{out}} = N_r - N_{r_{in}}$
query points outside the ROI, we index this embedding to create an ``Out-of-ROI-Embedding''
of shape $N_{r_{out}} \times 256$. The encoded occupancy and out-of-ROI-embedding are concatenated 
along the length dimension resulting in a tensor of shape $N_r \times 256$. 
An encoding of the depth of each query point along the ray is added to this tensor, which 
is processed by a sequence of CNN layers to combine information across the points.
Each CNN layer uses a kernel size of 4 and a stride of 2, and they have intermediate
dimensions of $64, 32, 16, 16, 16, 8$. ReLU activations are used between the CNN layers.
These 1D features are flattened (resulting in a feature vector per ray of dimensionality 232) and passed through an MLP using the ReLU activation function with intermediate dimensions
$232, 64, 32, 16, 1$ where the final
dimension is the scalar prediction for the depth of the ray $\mathcal{D}$.
All the lidar rays are used for training. 
We randomly sample a batch of 450 lidar rays (90 for each future lidar sweep) at each training step.

\subsection{Training Details}

We train \ourmodel{} for a total of 800,000 iterations across 16 GPUs (50,000 iterations each) with a batch size
of 1 on each GPU. 

\paragraph{Optimizer:}
We warmup the learning rate at a constant rate for the first 1000 iterations beginning from $8.0 \times 10^{-5}$
and ending at $8.0 \times 10^{-4}$. Then we use cosine schedule configured to bring the learning rate 
to $0$ by 50,000 iterations.
We use the AdamW optimizer \cite{kingma2014adam} with a weight decay of $1.0 \times 10^{-4}$.

\paragraph{Initialization:}
We follow \cite{agro2023implicit} in their initialization procedure for the linear layer that predicts the
attention offsets in the implicit decoder: we initialize the weights with a small standard deviation of $0.01$
and a bias of $0$.
This ensures that the initial offset prediction is small enough such that $\mathbf{r} = \mathbf{q}_{x,y} + \Delta\mathbf{q}_{x,y}$ still represents a point within the feature map $\mathbf{Z}$, but large enough such that the predicted offsets do not get stuck at $\Delta \mathbf{q}_{x,y} = \mathbf{0}$.

\paragraph{Training Data:}

For point cloud forecasting, we present results of \ourmodel{} on Argoverse 2, NuScenes, and KITTI. For each evaluation setting, we train a version of
\ourmodel{} on the corresponding training dataset split (for Argoverse 2 we use the Sensor dataset training split).

For BEV semantic occupancy forecasting, we train all models on the vehicle labels provided by the Argoverse 2 Sensor dataset.
We define the ``vehicle class'' to be the union of the Argoverse classes 
\texttt{REGULAR\_VEHICLE}, \texttt{LARGE\_VEHICLE}, \texttt{BUS}, \texttt{BOX\_TRUCK}, \texttt{TRUCK}, \texttt{VEHICULAR\_TRAILER},
\texttt{TRUCK\_CAB}, \texttt{SCHOOL\_BUS}, \texttt{ARTICULATED\_BUS}, \texttt{RAILED\_VEHICLE}, and \texttt{MESSAGE\_BOARD\_TRAILER}.

\section{Evaluation Details}

\subsection{Point Cloud Forecasting} \label{sec:point-cloud-forecasting-eval}

For point-cloud forecasting, we follow the evaluation procedure of \CMU{} \cite{khurana2023point}. The code is available online\footnote{
\href{https://github.com/tarashakhurana/4d-occ-forecasting}{https://github.com/tarashakhurana/4d-occ-forecasting}}, but we explain it in this section. 

\paragraph{Setting:}
On Argoverse 2 and KITTI, $H = 5$ past LiDAR sweeps are used as input at an interval of \SI{0.6}{s}, and the goal is to forecast point clouds
at 5 future timesteps $\{0.6, 1.2, \dots, 3.0\}\;s$.
On NuScenes we use $H = 6$ sweeps at an interval of $\SI{0.5}{s}$ as input, and forecast point clouds at 6 future timesteps $\{0.5, 1.0, \dots, 3.0\}\;s$.
Evaluation for Table 1. in the main paper are on the following dataset splits:
\begin{itemize}
    \item AV2: Argoverse 2 LiDAR dataset test split
    \item NuScenes: validation set\footnote{See \href{https://github.com/tarashakhurana/4d-occ-forecasting/issues/8}{https://github.com/tarashakhurana/4d-occ-forecasting/issues/8} for an explaination as to why the test set is not used.}
    \item KITTI: test set
\end{itemize}
We note that the Argoverse 2 point cloud forecasting challenge and the point-cloud forecasting results in Table. 2
of the main paper use the Argoverse 2 Sensor dataset test split.

\paragraph{Metrics:}
Point cloud evaluation has four metrics: Chamfer Distance (CD), Near Field Chamfer Distance (NFCD), depth L1 error, and depth relative L1 error (AbsRel),
each of which we explain below.

CD is computed as:
\begin{align}
    CD = \frac{1}{2N} \sum_{\mathbf{x} \in \mathbf{X}} \min_{\hat{\mathbf{x}} \in \mathbf{\hat{X}}} || \mathbf{x} - \hat{\mathbf{x}} ||^2_2 + \frac{1}{2M} \sum_{\hat{\mathbf{x}} \in \mathbf{\hat{X}}} \min_{\mathbf{x} \in \mathbf{X}} || \mathbf{x} - \hat{\mathbf{x}} ||^2_2,
\end{align}
where $\mathbf{X} \in \mathbb{R}^{N \times 3}$, $\hat{\mathbf{X}} \in \mathbb{R}^{M \times 3}$ represent the ground-truth and predicted point clouds, respectively.
NFCD is CD computed only on (ground truth and predicted) points inside the ROI of $[-70, 70]\;m$ in both the $x$ and $y$ axes, and $[-4.5, 4.5]\;m$ in the z axis around the ego vehicle.

L1 and AbsRel require the ground truth point-cloud $\mathbf{X}$ and the predicted point cloud $\hat{\mathbf{X}}$ to have a 1-1 correspondence of ray directions.
While for \ourmodel{} this is always the case because we predict depth along the rays from the future ground truth point clouds,
for some point-cloud forecasting methods this condition might not be met. Thus, for fair evaluation, \cite{khurana2023point} first fits a surface to the predicted point cloud,
and the intersection of the ground truth ray and that surface is computed to find the predicted ray depth. In practice, this is done by computing the projection of $\mathbf{X}$
and $\hat{\mathbf{X}}$ on the unit sphere, for each projected ground truth point finding the nearest projected forecasted point, and then setting the depth along that
ground truth ray equal to the depth of its nearest forecasted point. The result of this is a vector of 
forecasted depths along each LiDAR ray $\mathbf{\hat{D}} \in \mathbb{R}^{N}$ and ground truth depths ${\mathbf{D}} \in \mathbb{R}^{N}$.
Then L1 is calculated as
\begin{align}
    \text{L1} = \texttt{mean}(\texttt{abs}(\mathbf{D} - \mathbf{\hat{D}}))
\end{align}
and 
\begin{align}
    \text{AbsRel} = \texttt{mean}(\texttt{abs}( (\mathbf{D} - \hat{\mathbf{D}}) / \mathbf{D})),
\end{align}
where the vector division is element-wise.

\subsection{BEV Semantic Occupancy Forecasting}

\paragraph{Setting:}

We evaluate BEV semantic occupancy on the Argoverse 2 Sensor validation dataset. The LiDAR input is $H = 5$ past LiDAR sweeps at an interval of \SI{0.6}{s} to match
the unsupervised pre-training of \ourmodel{}, and the BEV semantic occupancy predictions are evaluated on 2D grids of points
of size \SI{140}{m} by \SI{140}{m} with a spatial resolution of \SI{0.2}{m} centered on the ego at future timesteps $\{0, 0.6, 1.2, \dots, 3.0\}\;s$.

\paragraph{Metric:}

Mean Average Prediction (mAP) is computed on (a) the predicted occupancy values from all the grid points across all evaluation examples, and (b) the corresponding
ground truth occupancy values at those points. For thresholds $[0, 0.01, 0.02, \dots, 0.99, 1.0]$, the number of true positives (TP),
false positives (FP), and false negatives (FN), are calculated. At each threshold value, we obtain a precision and recall
\begin{align}
    \text{precision} = \frac{TP}{TP + FP}, \; \text{recall} = \frac{TP}{TP + FN},
\end{align}
and mAP is then computed as the area under the curve formed by precision (y-axis) and recall (x-axis) across all thresholds.

\section{Baseline Implementation Details}

\paragraph{\freespacerendering{}:}

We use same model architecture as \ourmodel{}, described in \cref{sec:implementation-details}.
Mirroring the training of the learned renderer (see \cref{sec:rendering-header}),
we sample $450$ future lidar rays within the \SI{3}{s} time horizon.
Considering a single lidar ray, we generate $N_r = 2000$ evenly spaced query points along the lidar ray
at a step size of $\epsilon_r = \SI{0.1}{m}$ from the emitting sensor.
The model is queried at these points to produce occupancy predictions
$\hat{\mathbf{o}} = [\hat{o}_1, \dots, \hat{o}_{N_r}] \in [0, 1]^{N_r}$
along each the ray.
Following \cite{khurana2023point}, we complement the cumulative maximum of the occupancy along the LiDAR ray
to obtain a prediction of the free-space:
\begin{align}
    \hat{\mathbf{v}} = 1 - \texttt{cummax}(\hat{\mathbf{o}}).
\end{align}
Using the ground-truth future lidar point we can generate a free-space label
$\mathbf{v} =  [v_1, \dots, v_{N_r}] \in \{0, 1\}^{N_r}$ which is $1$ for all points along the ray
prior to the LiDAR point and $0$ for all points along the ray after the LiDAR point.
Then the model is trained with cross entropy loss between the free-space label and the free-space prediction.

\paragraph{\depthrendering{}:}

We use the same model architecture as \ourmodel{}, described in \cref{sec:implementation-details}.
Similarly to \freespacerendering{}, we sample future lidar rays and query the model along each lidar ray to produce
occupancy predictions
$\hat{\mathbf{o}} = [\hat{o}_1, \dots, \hat{o}_{N_r}] \in [0, 1]^{N_r}$. 
The expeceted depth along the ray is computed as
\begin{align}
    \mathbb{E}[\ell] = \sum_{i = 0}^{N_r} (\epsilon_r * i) \left( \left( \prod_{j = 0}^{i-1} (1 - \hat{o}_j) \right) \hat{o}_i \right),
\end{align}
and the loss against the ground truth LiDAR point depth $\ell_{gt}$ is computed as $|\ell_{gt} - \mathbb{E}[\ell]|$, averaged
across all rays in the batch.

\paragraph{MP3}

We use the same encoder as \ourmodel{} for a fair comparison (see \cref{sec:implementation-details}). 
MP3 decodes initial occupancy and forward flow over time using a fully convolutional header, and the future
occupancy is obtained by warping the initial occupancy with the temporal flow as described 
in the original paper \cite{casas2021mp3}.

\paragraph{ImplicitO}

We use the same model architecture as \ourmodel{} (which is very similar to the orignal paper \cite{agro2023implicit}, see \cref{sec:implementation-details}),
but the query point feature size is 3 $(x, y, t)$ because they are in BEV instead of 3D.

\section{Additional Quantitative Results}

\subsection{Effect of Freezing the Encoder During Fine-tuning}

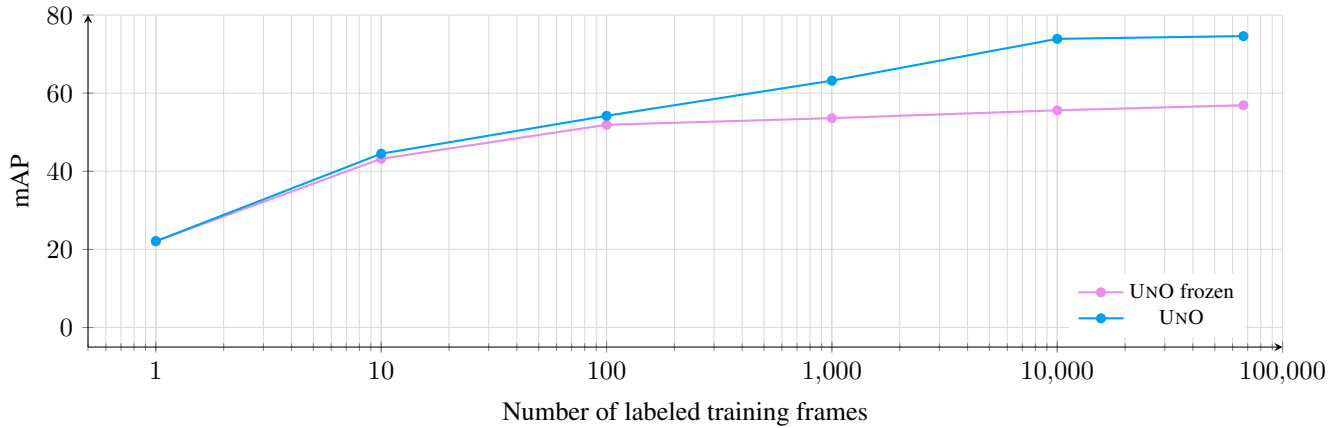
\begin{figure}
    \centering
    \vspace{-5pt}
    \begin{tikzpicture}
    \definecolor{clr1}{HTML}{ef9b20}
    \definecolor{clr2}{HTML}{bdcf32}
    \definecolor{clr3}{RGB}{0,160,236}
    \definecolor{clr4}{HTML}{E88EED}
    \begin{axis}[
        xlabel=Number of labeled training frames,
        ylabel=mAP,
        xmode=log,
        log ticks with fixed point,
        legend pos=south east,
        legend style={nodes={scale=0.8, transform shape}, draw=none}, 
        width=\linewidth,
        height=6cm,
        axis lines=left,
        xmin=0.5,
        xmax=100000,
        ymin=-5,
        ymax=80,
        grid=both,
        grid style={line width=.1pt, draw=gray!30},
    ]

    \addplot[color=clr4, mark=*, mark size=1.5pt, thick] coordinates {
        (1, 22.1)
        (10, 43.2)
        (100, 51.9)
        (1000, 53.6)
        (10000, 55.6)
        (67000, 56.9)
    };
    \addlegendentry{\ourmodel{} frozen}

    \addplot[color=clr3, mark=*, mark size=1.5pt, thick] coordinates {
        (1, 22.1)
        (10, 44.5) %
        (100, 54.2) %
        (1000, 63.2) %
        (10000, 73.9) %
        (67000, 74.6) %
    };
    \addlegendentry{\ourmodel{}}

    \end{axis}
    \end{tikzpicture}
    \caption{
        BEV semantic occupancy results comparing \ourmodel{} Frozen, which fine-tunes only the occupancy decoder, to
        \ourmodel{}, which fine-tunes all parameters.
    }
    \label{fig:supp-finetuning}
\end{figure}

In this experiment, we want to investigate how fine-tuning \ourmodel{} on BEV semantic occupancy is affected
by amount of model parameters that are updated.
To do this, we fine-tune \ourmodel{} on the task of BEV semantic occupancy forecasting, but freeze all the parameters
in the LiDAR encoder (see \cref{sec:implementation-details}), which we call \ourmodel{} Frozen. \cref{fig:supp-finetuning} presents the results across
multiple levels of supervision, against \ourmodel{} with all parameters fine-tuned.
We notice that for up to 100 labelled frames, the performance of \ourmodel{} Frozen is close to \ourmodel{}, but their performance
diverges as the amount of labelled data increases. This is likely because fine-tuning only the occupancy decoder, which has very few parameters (0.06M out of a total of 17.4M),
has limited representational power in changing the output fully from geometric occupancy (which it was pre-trained for) to BEV semantic occupancy.

\subsection{Effect of Learned Renderer:}

\begin{table}[]
    \setlength\tabcolsep{4pt} %
    \centering
    \footnotesize
    \begin{tabular}{lllll}
    \toprule
     Rendering     & \begin{tabular}[c]{@{}l@{}}NFCD\\ (\SI{}{\meter\squared}) $\downarrow$\end{tabular} & \begin{tabular}[c]{@{}l@{}}CD\\ (\SI{}{\meter\squared}) $\downarrow$\end{tabular} & \begin{tabular}[c]{@{}l@{}}AbsRel\\ (\SI{}{\percent}) $\downarrow$\end{tabular} & \begin{tabular}[c]{@{}l@{}}L1\\ (\SI{}{\meter\squared}) $\downarrow$\end{tabular} \\ \midrule
                Threshold 0.5       & 3.09                                                                                & 22.75                                                                             & 13.34                                                                           & 3.80                                                                              \\ %
                Threshold 0.8       & 2.09                                                                                & 18.49                                                                             & 11.33                                                                           & 3.01                                                                              \\ %
                Threshold 0.9       & 1.63                                                                                & 18.16                                                                             & 11.16                                                                           & 2.67                                                                              \\ %
                Threshold 0.95      & 1.48                                                                                & 23.02                                                                             & 14.90                                                                           & 2.91                                                                              \\ %
                Learned             & \textbf{0.83}                                                                       & \textbf{8.10}                                                                     & \textbf{10.09}                                                                  & \textbf{2.03}                                                                     \\ \bottomrule %
    \end{tabular}
    \vspace{3pt}
    \caption{Comparing point cloud prediction performance of \ourmodel{} with different rendering methods (thresholding occupancy and learned rendering) on the Argoverse 2 Sensor dataset.}
    \label{tab:ths-vs-learned}
    \vspace{-0.4cm}
\end{table}

In the main paper we described that generating point cloud forecasts from
occupancy is a non-trivial task, motivating our use of the learned renderer.
In this experiment, we compare it to a simple baseline that thresholds the forecasted occupancy values.
Specifically, to find the depth of a ray, we walk along the ray direction starting
from the sensor location, and stop when the occupancy value exceeds a given threshold.
\cref{tab:ths-vs-learned} presents the results of this comparison on the Argoverse 2 Sensor dataset,
where we see large improvements from using the learned rendering method over the thresholding method with
any threshold.
This is expected since the learned renderer can better contextualize the scene by taking into account
a set of occupancy values predicted along the ray.

\subsection{Supervised vs Unsupervised Occupancy}

In this section we compare the unsupervised 4D geometric occupancy predictions
to supervised 4D semantic occupancy predictions (trained using bounding box labels).
To do this fairly across geometric and semantic occupancy predictions,
we employ our recall metric described in the main paper, which ignores background regions (which geometric occupancy includes, but semantic occupancy does not)
and focuses on foreground actors and free-space (which both forms of occupancy should understand). 

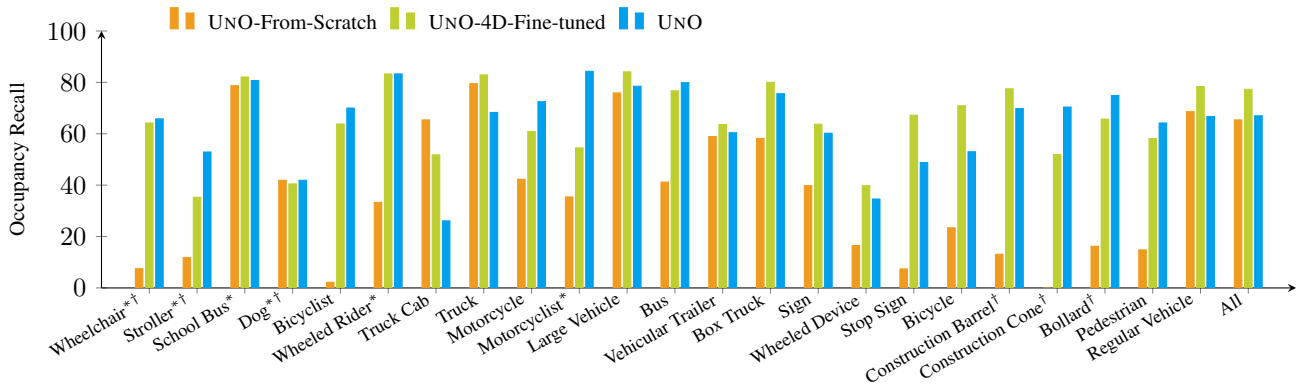
\begin{figure*}
    \centering
    \begin{tikzpicture}
        \definecolor{clr1}{HTML}{ef9b20}
        \definecolor{clr2}{HTML}{bdcf32}
        \definecolor{clr3}{RGB}{0,160,236}
        \definecolor{clr4}{HTML}{E88EED}
        \definecolor{clr5}{RGB}{221,194,240}
        \begin{axis}[
            ybar,
            bar width=0.1cm,
            enlarge x limits=0.05,
            ylabel={\footnotesize{Occupancy Recall}},
            width=\textwidth,
            height=5cm,
            xtick=data,
            ybar=1pt, %
            xmin=3,
            xmax=28,
            ymin=0,
            ymax=100,
            axis lines=left,
            xticklabels={
                \scriptsize Wheelchair$^{*\dagger}$, %
                \scriptsize Stroller$^{*\dagger}$, %
                \scriptsize School Bus$^*$, %
                \scriptsize Dog$^{*\dagger}$, %
                \scriptsize Bicyclist, %
                \scriptsize Wheeled Rider$^*$, %
                \scriptsize Truck Cab, %
                \scriptsize Truck, %
                \scriptsize Motorcycle, %
                \scriptsize Motorcyclist$^*$, %
                \scriptsize Large Vehicle, %
                \scriptsize Bus, %
                \scriptsize Vehicular Trailer, %
                \scriptsize Box Truck, %
                \scriptsize Sign, %
                \scriptsize Wheeled Device, %
                \scriptsize Stop Sign, %
                \scriptsize Bicycle, %
                \scriptsize Construction Barrel$^\dagger$, %
                \scriptsize Construction Cone$^\dagger$, %
                \scriptsize Bollard$^\dagger$, %
                \scriptsize Pedestrian, %
                \scriptsize Regular Vehicle, %
                \scriptsize All, %
            },
            x tick label style={rotate=30,anchor=east},
            legend style={at={(0.05, 0.95)},anchor=south west, legend columns=-1, column sep=3pt, draw=none},
            ]

            \addplot[fill=clr1, color=clr1] coordinates {
                (4,  7.5)
                (5,  11.8)
                (6,  78.7)
                (7,  41.9)
                (8,  2.2)
                (9,  33.3)
                (10, 65.4)
                (11, 79.5)
                (12, 42.3)
                (13, 35.4)
                (14, 75.9)
                (15, 41.2)
                (16, 58.9)
                (17, 58.2)
                (18, 39.8)
                (19, 16.5)
                (20, 7.4)
                (21, 23.4)
                (22, 13.1)
                (23, 0.0)
                (24, 16.2)
                (25, 14.8)
                (26, 68.6)
                (27, 65.4)
            };
            \addlegendentry{\footnotesize{\ourmodel{}-From-Scratch}}

            \addplot[fill=clr2, color=clr2] coordinates {
                (4,  64.2)
                (5,  35.3)
                (6,  82.1)
                (7,  40.5)
                (8,  63.8)
                (9,  83.3)
                (10, 51.8)
                (11, 82.9)
                (12, 60.9)
                (13, 54.5)
                (14, 84.1)
                (15, 76.7)
                (16, 63.6)
                (17, 80.0)
                (18, 63.7)
                (19, 39.8)
                (20, 67.2)
                (21, 70.9)
                (22, 77.5)
                (23, 51.9)
                (24, 65.7)
                (25, 58.1)
                (26, 78.4)
                (27, 77.3)
            };
            \addlegendentry{\footnotesize{\ourmodel{}-4D-Fine-tuned}}

            \addplot[fill=clr3, color=clr3] coordinates {
                (4,  65.8)
                (5,  52.9)
                (6,  80.7)
                (7,  41.9)
                (8,  70.0)
                (9,  83.3)
                (10, 26.1)
                (11, 68.3)
                (12, 72.5)
                (13, 84.3)
                (14, 78.5)
                (15, 79.9)
                (16, 60.4)
                (17, 75.6)
                (18, 60.2)
                (19, 34.6)
                (20, 48.8)
                (21, 53.0)
                (22, 69.8)
                (23, 70.4)
                (24, 74.9)
                (25, 64.2)
                (26, 66.7)
                (27, 67.0)
            };
            \addlegendentry{\footnotesize{\ourmodel{}}}

        \end{axis}
    \end{tikzpicture}
    \caption{
        Recall of \ourmodel{} compared to other supervised and semi-supervised baselines on the Argoverse 2 Sensor dataset, averaged across the prediction horizon. Recall was computed at a target precision of 0.7.
        $^*$ denotes the rarest $25\%$ of classes, and $^\dagger$ denotes the smallest (by average bounding box volume) $25\%$ of classes.
    }
    \label{fig:classes-bar-plot-supervised-unsupervised}
    \vspace{-1em}
\end{figure*}

In \cref{fig:classes-bar-plot-supervised-unsupervised}, we evaluate the multi-class 3D occupancy recall
of:
\begin{enumerate}
    \item \ourmodel{}: Unsupervised 4D geometric occupancy pre-training only (described in main paper).
    \item \ourmodel{}-From-Scratch: Using the \ourmodel{} architecture, but with no unsupervised pre-training. Instead,
    this model is directly trained only on labeled bounding box data to forecast 4D semantic occupancy. We group
    all actor classes into a single class.
    \item \ourmodel{}-4D-Fine-Tuned: \ourmodel{} with 4D geometric occupancy pre-training and additional fine-tuning to forecast 4D semantic occupancy
    (grouping all actors into a single class). 
\end{enumerate}
For this fine-tuning process, we train both the supervised and the fine-tuned models with all available data. 

Remarkably, \ourmodel{} (which uses no labeled data) achieves overall better recall
than \ourmodel{}-From-Scratch. We hypothesize this is because the supervised model has 
to learn to classify between foreground and background objects, which can make it hard to learn
the occupancy of rare classes, small classes, or classes that are hard to distinguish from the background.
This is highly relevant to safe autonomous driving for detecting potentially unknown or rarely seen 
road agents.

\ourmodel{}-4D-Fine-Tuned slightly improves recall (see the ``All" class) over \ourmodel{} and greatly improves over \ourmodel{}-From-Scratch.
This highlights the expressive representations learned during the \ourmodel{} pre-training procedure.

\subsection{Ablation of the $\delta$ Parameter}

$\delta = \SI{0.1}{m}$ was used for all three datasets, however, our pre-training procedure is quite robust to the choice of $\delta$.
Below we ablate the choice of $\delta$ on all tasks. First, we note that all the possible values below result in SOTA performance.
Second, it shows that increasing $\delta$ from \SI{0.1}{m} to \SI{1.0}{m} indeed degrades performance.
Qualitatively, we notice very small differences in occupancy forecasts,
with a slight enlargement of shapes with $\delta = \SI{1.0}{m}$ that is apparent in the tree trunks.

\begin{figure}
    \centering
    \begin{tabular}{llllllll}
    \toprule 
                                                                                & \multicolumn{4}{c}{\textbf{Point Cloud Forecasting}}                                                                                                                                                                                                                                                                                          & \multicolumn{2}{c}{\begin{tabular}[c]{@{}l@{}}\textbf{BEV Semantic}\\\textbf{Occupancy}\end{tabular}}              & \begin{tabular}[c]{@{}l@{}}\textbf{4D}\\\textbf{Occupancy}\end{tabular} \\
                                                                                \cmidrule(l{5pt}r{5pt}){2-5} \cmidrule(l{5pt}){6-7} \cmidrule(l{5pt}){7-8}
    \begin{tabular}[c]{@{}l@{}}$\delta\;(\text{m})$\end{tabular} & \begin{tabular}[c]{@{}l@{}}NFCD\\ (\SI{}{\meter\squared}) $\downarrow$\end{tabular} & \begin{tabular}[c]{@{}l@{}}CD\\ (\SI{}{\meter\squared}) $\downarrow$\end{tabular} & \begin{tabular}[c]{@{}l@{}}AbsRel\\ (\SI{}{\percent}) $\downarrow$\end{tabular} & \begin{tabular}[c]{@{}l@{}}L1\\ (\SI{}{\meter\squared}) $\downarrow$\end{tabular} & \begin{tabular}[c]{@{}l@{}}mAP\\$\uparrow$\end{tabular}      & \begin{tabular}[c]{@{}l@{}}Soft-IoU\\$\uparrow$\end{tabular}        &  Recall @ 0.7                 \\ \midrule
    1.00                                                     & 0.85                                                                                & 8.22                                                                                  & 11.5                                                                            & 2.08                                                                              & 50.3                                                         & 21.1                                                                &  65.7                         \\ %
    0.01                                                     & \textbf{0.82}                                                                       & \textbf{8.06}                                                                         & 10.8                                                                            & 2.04                                                                              & 52.0                                                         & 22.2                                                                &  65.8                         \\  %
    0.10                                                     & 0.83                                                                                & 8.10                                                                                  & \textbf{10.1}                                                                   & \textbf{2.03}                                                                     & \textbf{52.3}                                                & \textbf{22.3}                                                       &  \textbf{67.0}                         \\ \bottomrule %
    \end{tabular}
    \caption{Ablation of the $\delta$ parameter used in the training of \ourmodel{}.} \label{tab:delta-ablation}
\end{figure}

\section{Additional Qualitative Results}

\subsection{Occupancy and Point Cloud Forecasting} \label{sec:occ-qualitative}

\cref{fig:supp-qualitative-1,fig:supp-qualitative-2,fig:supp-qualitative-3}
show the geometric occupancy forecasts and point cloud forecasts of \ourmodel{}
and various baselines from the main paper. A few overarching themes are:
\begin{itemize}
    \item \CMU{} struggles to abstract point clouds into occupancy; it mainly predicts occupancy only where there
    are observed points.
    \item \CMU{} and \depthrendering{} don't model motion and object extent.
    \item \freespacerendering{} understands extent, but does not model motion very accurately.
    \item \unbalancedourmodel{} is under-confident (even on static background areas we see relatively low confidence).
    It has reasonable motion predictions but without much multi-modality, and it is worse at understanding of extent than \ourmodel{}. 
    \item \ourmodel{} models motion, extent, and multi-modality. 
\end{itemize}
We use the following numbers to highlight interesting things in \cref{fig:supp-qualitative-1,fig:supp-qualitative-2,fig:supp-qualitative-3}.
\begin{enumerate}
    \item Struggling to abstract point clouds into occupancy.
    \item Inaccurate predictions of extent.
    \item Predicting a moving actor stays static.
    \item Disappearing occupancy on an actor.
    \item Inaccurate motion predictions.
    \item Under-confidence on background areas.
    \item Under-confidence on actors.
    \item Multi-modal predictions.
    \item Accurate motion predictions for non-linear actors (e.g., turning, lane-changing).
\end{enumerate}
\FloatBarrier %
\input{figures/supp_scene_qualitative_1.tex}
\FloatBarrier
\input{figures/supp_scene_qualitative_2.tex}
\FloatBarrier
\input{figures/supp_scene_qualitative_3.tex}
\FloatBarrier

\subsubsection{Occupancy Perception} \label{sec:occ-perception}

To contextualize \ourmodel{}'s occupancy predictions, we visualize them along side camera images provided in Argoverse 2 in \cref{fig:camera}.
We observe that \ourmodel{} is able to perceive all objects in the scene, including relatively rare occurrences like construction, over a large region of interest.

\FloatBarrier
\input{figures/camera.tex}
\FloatBarrier

\subsubsection{Failure Cases} 

The main failure cases of \ourmodel{} are due to the limited range and noise of LiDAR sensors, resulting in limited supervision.
\Cref{fig:failure} shows how this can manifest in uncertain occupancy far from the ego (a),
and also at points high above the ego (b) since the LiDAR rays never return from the sky.
\Cref{fig:camera}
also shows that there is room for improvement in capturing fine details like the shape of the cyclist in Scene 1.
Finally, we note that the long occupancy ``tails" behind certain vehicles are not necessarily a failure case.
These are expected when predicting marginalized occupancy probabilities on multi-modal scenarios where a vehicle could stay still or accelerate.

\FloatBarrier
\input{figures/failure_cases.tex}
\FloatBarrier

\end{document}